\documentclass[11pt]{article}

\usepackage[final]{acl}

\usepackage{times}
\usepackage{latexsym}
\usepackage[T1]{fontenc}
\usepackage[utf8]{inputenc}
\usepackage{microtype}
\usepackage{inconsolata}
\usepackage{graphicx}
\usepackage{booktabs}
\usepackage{multirow}
\usepackage{xcolor}
\usepackage{amsmath}
\usepackage{amssymb}
\usepackage{listings}
\usepackage{algorithm}
\usepackage{algorithmic}
\usepackage{tikz}
\usetikzlibrary{shapes.geometric,arrows.meta,positioning,fit,calc,backgrounds}

\title{Compile, Don't Memorize: A Context Compilation Architecture (CCA) for In-Context Learning}

\author{
  \textbf{Jinhu Qi}\textsuperscript{1} \quad \textbf{Minda Hu}\textsuperscript{1} \quad \textbf{Wentao Zhang}\textsuperscript{2} \quad \textbf{Weiqiang Jin}\textsuperscript{3} \\
  \textbf{Yanyu Chen}\textsuperscript{1} \quad \textbf{Junli Wang}\textsuperscript{4} \quad \textbf{Irwin King}\textsuperscript{1} \\
  \textsuperscript{1}The Chinese University of Hong Kong \quad \textsuperscript{2}Macao Polytechnic University \\
  \textsuperscript{3}Xi'an Jiaotong University \quad \textsuperscript{4}University of Science and Technology of China \\
  \footnotesize \texttt{\{jhqi25, mdhu22, yychen25, king\}@cse.cuhk.edu.hk} \ \ \texttt{p2522808@mpu.edu.mo} \\
  \footnotesize \texttt{weiqiangjin@stu.xjtu.edu.cn} \ \ \texttt{junliwang@mail.ustc.edu.cn}
}

\begin{document}
\maketitle

\begin{abstract}
Large language models (LLMs) increasingly handle in-context learning (ICL) tasks where a long, novel context defines the rules, knowledge, and output schema for a series of questions. On benchmarks that grade against every detail of the context, even strong open-weights models pass only 12--16\% of tasks: a single overlooked rule fails the whole response. We argue this brittleness is structural: the dominant ``read-and-reason'' paradigm asks the model to extract, plan, generate, and self-verify in one forward pass. We therefore ask whether explicit context compilation can fix it, how it compares to existing long-context strategies (gist retrieval, multi-agent self-play), and where the resulting harness benefit holds across task structure and model scale. We propose the \textbf{Context Compilation Architecture (CCA)}, whose central novelty is a \emph{typed intermediate representation (IR) with fixed slots} (\texttt{rules.\{must\_do, must\_not, conditional\}}, \texttt{output\_spec}, \texttt{available\_tools}, \texttt{data\_profile}) into which any prose context is compiled once; executable verifiers and a violation-gated correction loop follow as downstream consequences. On CL-bench (1{,}899 tasks across 4 open base models), CCA outperforms vanilla prompting and two long-context baselines (ReadAgent-P, Ctx2Skill) on every base model, lifting Kimi K2.5 from \textbf{15.4\% to 21.4\%} with gains concentrated on rule-dense sub-categories. Code and cached completions are available at \url{https://github.com/TonyQJH/cca-emnlp2026}.
\end{abstract}

\section{Introduction}
\label{sec:intro}

Modern LLMs are routinely deployed in scenarios where a single conversation contains tens to hundreds of thousands of characters of novel context (domain documentation, multi-agent workflows, custom DSLs, complex business rules), and the model is asked to comply with this context across many downstream questions. Benchmarks designed for this setting, such as CL-bench~\citep{clbench}, grade responses against detailed \emph{rubrics}: each of 5--20 criteria per question must be met, and a single missed constraint marks the task as failed. Empirically, even strong open base models pass only 12--16\% of such tasks under vanilla prompting (Table~\ref{tab:main-results}).

We hypothesize that the issue is not raw reasoning capability but a saturation of the read-and-reason pipeline: any \emph{single} constraint is usually within reach, yet the model is asked to perform extraction, planning, generation, and self-verification simultaneously in a single forward pass, and small details fall through the cracks. As context length grows and rubrics multiply, the joint probability of catching every required element~collapses.

We draw inspiration from \textbf{movable type printing}. Before Bi Sheng's 11th-century invention, reproducing a new page required carving an entire fixed wooden plate that was expensive, inflexible, and single-use. Movable type changed this by introducing a two-stage workflow: a \emph{compositor} first reads the new manuscript and selects individual type pieces into a custom plate; only then does the press mechanically reproduce the page. We argue current ICL methods resemble fixed-plate printing: the model carries pre-trained ``plates'' (memorized procedures) and tries to adapt them by force to whatever context arrives. Instead, we propose to \emph{compile} the context: treat the raw input as a manuscript, extract the typed pieces (rules, entities, exact terms, output schema), and assemble them into an executable plate that mechanically verifies and produces the answer.

Concretely, we introduce the \textbf{Context Compilation Architecture (CCA)}, whose central novelty is a \emph{typed context representation}: the compiled IR --- a fixed-slot JSON vocabulary --- is the representational contract the rest of the pipeline consumes, and executable per-context verifiers plus a violation-gated correction step follow as downstream consequences, not co-equal contributions. Where Self-Refine~\citep{madaan2023selfrefine}, Ctx2Skill~\citep{si2026ctx2skill}, and ReadAgent-P~\citep{lee2024readagent} operate on prose throughout, this representational shift turns the ``did I remember everything?'' question into the mechanical ``did my code-verifier pass?''.

CCA is, in modern terms, a concrete instance of \emph{harness engineering} \citep{bockeler2026harness}: the emerging discipline that treats the LLM as a frozen reasoning utility and derives production reliability from the surrounding~pipeline.

\paragraph{Running example.} Task \texttt{54ff6369} (14 rubrics): Compiler yields 50 rules, CodeGen synthesizes verifiers, Reasoner-1 drafts, 4 violations flagged, Reasoner-2 fixes only those---CCA 14/14 vs vanilla fail.

\paragraph{Research questions.} We organize the paper around three questions any context-compilation harness must answer. \textbf{RQ1 (Effectiveness):} does explicitly compiling context into a structured IR with executable verifiers improve pass rate on rubric-graded ICL relative to vanilla prompting, and which CCA components are responsible for the improvement? \textbf{RQ2 (Comparison with existing strategies):} how does explicit context compilation compare to gist-memory retrieval (ReadAgent-P) and self-play skill libraries (Ctx2Skill) under rubric-based grading, and which design properties of each method are well- or poorly-matched to this evaluation regime? \textbf{RQ3 (Where and for whom):} where does the compilation harness help most, and how does the benefit scale with the underlying model's capacity and with the task's structural~density?

\paragraph{Contributions.} (RQ1) We introduce CCA, a context-driven compilation pipeline that auto-generates per-context verification tooling and improves rubric-graded ICL pass rate over vanilla prompting on every base model tested (§\ref{sec:method},~§\ref{sec:results}); we localise the load-bearing component via a controlled ablation (§\ref{sec:rq1-answer}). (RQ2) We characterise how two strong long-context strategies (gist-memory retrieval and multi-agent self-play) interact with rubric-based grading and identify design properties well- or poorly-matched to this regime (§\ref{sec:rq2-answer}). (RQ3) We isolate the task-structural and model-capacity factors that bound the harness benefit, through per-domain and per-sub-category breakdowns~(§\ref{sec:rq3-answer}).

\section{Related Work}
\label{sec:related}

\textbf{Long-context in-context learning.} ReadAgent-P paginates a context, stores per-page gists, and at query time looks up and re-fetches relevant pages \citep{lee2024readagent}; this suits long-document QA where information is concentrated, but its gist step omits verbatim facts that rubric grading requires. Ctx2Skill casts context learning as skill augmentation via a Challenger/Reasoner/Judge/Proposer/Generator self-play loop producing a natural-language skill library retrieved at inference time \citep{si2026ctx2skill}, with AutoSkill a related single-stage variant \citep{yang2026autoskill}; the resulting library encodes procedural guidance rather than per-criterion obligations. Broader work combines retrieval, routing, and compression: RAG and variants vary retrieval granularity, order, or routing \citep{lewis2020rag,jiang2024longrag,yu2024defense,li2024selfroute}; LLMLingua-2 and ICAE compress prompts or encode memory slots \citep{pan2024llmlingua2,ge2024icae}; Chain-of-Agents distributes inputs across cooperating readers \citep{zhang2024chainagents}. Empirical studies further show that even within nominal windows long-context models struggle with mid-context evidence \citep{liu2024lostmiddle} and that effective context lengths often fall short of claimed windows \citep{hsieh2024ruler}, motivating explicit structure extraction over positional access. CCA differs by compiling each context once into a typed IR that \emph{preserves} exact terms and into executable verifier code that mechanically checks each constraint.

\textbf{Code-as-reasoning and verification.} Building on chain-of-thought prompting \citep{wei2022cot}, PAL, PoT, and Chain-of-Code translate a problem into executable programs, delegating computation to an interpreter \citep{gao2023pal,chen2023pot,li2024coc}. LLM-as-judge made model-based evaluation a standard self-checking interface \citep{zheng2023judge}, with extensions to iterative feedback, verbal-feedback retry, tool-interactive critiquing, reasoning checks, step-level verifier benchmarks, and synthesised Python verifier sets \citep{madaan2023selfrefine,shinn2023reflexion,gou2024critic,miao2024selfcheck,jacovi2024weakest,pezeshkpour2026autopyverifier}. In all of these, code is an answer-producing computation generated per question; in CCA, code is a per-context verifier generated once and reused to check whether a natural-language answer satisfies the context's typed rules, not to produce the answer.

\textbf{Harness engineering and agent infrastructure.} Practitioner discourse frames harness engineering as the execution environment, tool interfaces, contextual guides, lifecycle, observability, verification, and governance surrounding a frozen model (the ETCLOVG taxonomy) \citep{bockeler2026harness,sylph2026harness,li2026agentsurvey}, arguing that the gap between production-ready agents and the $\sim$$88\%$ that never ship lies in this layer rather than in raw model capability \citep{bockeler2026harness}. Adjacent infrastructure work supplies reusable agent abstractions (AutoGen, AgentScope, MetaGPT, SWE-agent) \citep{wu2024autogen,gao2024agentscope,hong2024metagpt,yang2024sweagent}. Closest in spirit, DSPy \citep{khattab2023dspy} compiles declarative LM pipelines into self-improving programs by optimizing prompts and weights across a fixed program structure; CCA differs in unit of compilation, compiling each given context once into a typed IR and per-context Python verifiers rather than a fixed pipeline of LM calls. To our knowledge, CCA is the first method to treat in-context learning itself as a \emph{compilation} problem (producing typed structure and executable verifier code from each given context), and to position the resulting pipeline within the harness-engineering~discipline.
\section{Context Compilation Architecture (CCA)}
\label{sec:method}

\subsection{Overview}

\begin{figure*}[t]
\centering
\includegraphics[width=\textwidth]{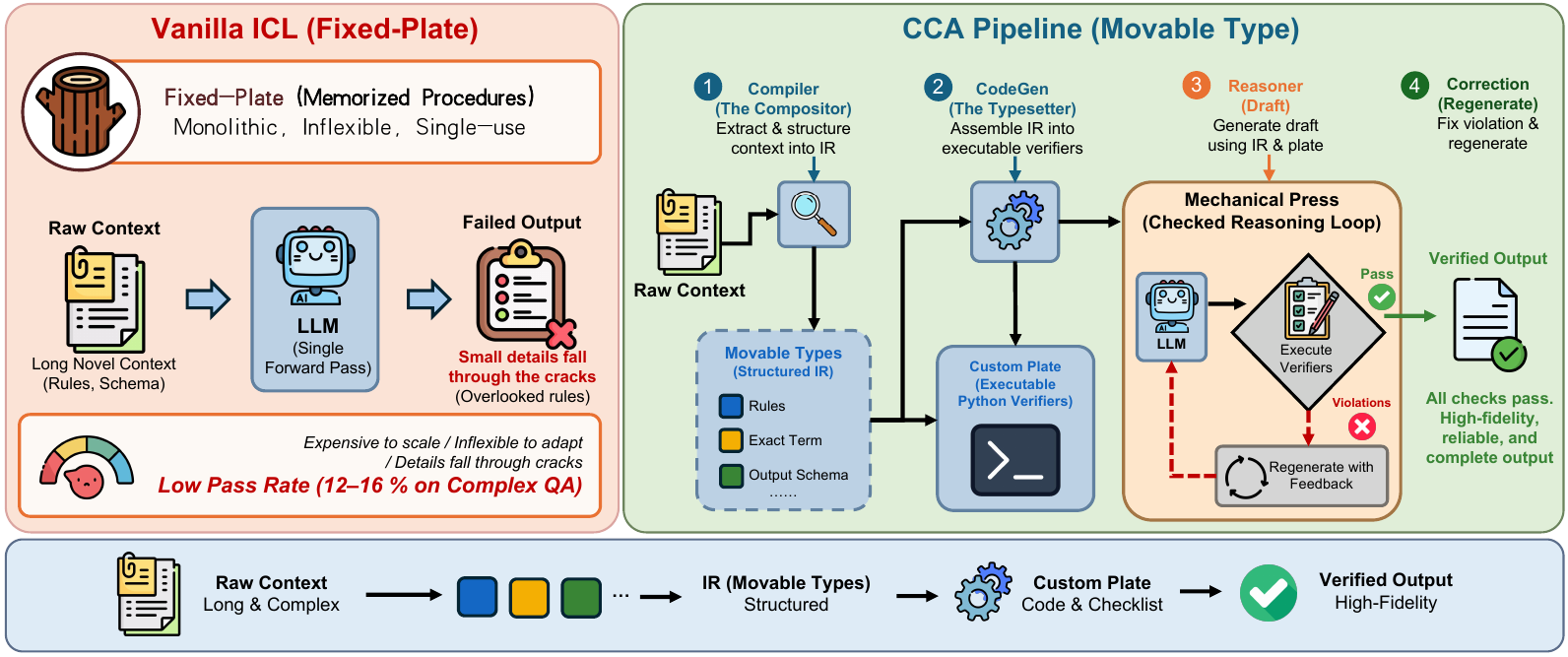}
\caption{\textbf{CCA pipeline vs vanilla ICL.}
\textbf{Left} (red, ``Fixed-Plate''): vanilla ICL pushes the long context through a single LLM forward pass; small details fall through the cracks and pass rate stays at $12$--$16\%$ on rubric-graded benchmarks (Table~\ref{tab:main-results}).
\textbf{Right} (green, ``Movable Type''): CCA decomposes the context into reusable typed pieces in four stages---(1) \emph{Compiler} (\emph{the compositor}) extracts a JSON IR of rules, exact terms, workflow, and output schema (F1); (2) \emph{CodeGen} (\emph{the typesetter}) assembles up to three Python verifiers from the IR (F3, F5); (3) \emph{Reasoner} drafts an answer using the IR as an explicit checklist (F2); (4) \emph{Correction} regenerates with violation feedback when verifiers flag $\geq 2$ violations (F4, F6). Per-context artifacts (stages 1--2) are amortized over every question sharing the context; per-task stages (3--4) run once per question. Head/tail context truncation (F7) is described in §\ref{sec:method}. The pipeline converts the diffuse self-judgment ``did I remember everything?'' into the mechanical check ``did my code-verifier pass?''.}
\label{fig:architecture}
\end{figure*}

CCA processes a single \emph{context group} (a context document plus one or more downstream tasks sharing that context, 1--12 tasks per context in CL-bench) in four stages (Figure~\ref{fig:architecture}). Stages 1--2 (Compiler, CodeGen) run \textbf{once per context}; stages 3--4 (Reasoner-1; Verifier Execution + Reasoner-2 correction) run \textbf{once per task}.

\subsection{Pipeline as a Composition}
\label{sec:pipeline-formalism}

CCA's pipeline mirrors its two-phase structure in Figure~\ref{fig:architecture}: a \textbf{\emph{per-context}} phase compiles a context $c$ once into reusable artifacts, and a \textbf{\emph{per-task}} phase reuses those artifacts to draft and conditionally correct one answer per downstream task. Given a raw context $c$, its task set $\mathcal{T}_c$ (each $t \in \mathcal{T}_c$ carrying a question $q_t$), and the instruction prompts $\mathcal{I}$ shared across the four LLM stages, the joint probability of all final outputs factorises as:

{\small
\begin{align}
&P\big(\{y_t\}_{t \in \mathcal{T}_c} | c, \mathcal{I}\big) = \underbrace{P(I_c | c, \mathcal{I}) \cdot P(\mathcal{M}_c | I_c) \cdot P(s_c | \mathcal{M}_c, c)}_{\textit{Per-context}\,(\text{F1, F3, F5})} \notag\\
&\cdot \underbrace{\prod_{t \in \mathcal{T}_c} P(d_t | I_c, s_c, \tau(c), q_t) \cdot P(y_t | d_t, \mathcal{V}_c, \theta)}_{\textit{Per-task}\,(\text{F2, F4, F6, F7})}
\label{eq:cca}
\end{align}
}

\noindent where $I_c$ is the typed JSON IR emitted by the Compiler ($\mathcal{C}$); $\mathcal{M}_c \subseteq \{\mathtt{RC}, \mathtt{FV}, \mathtt{DA}\}$ is the verifier module set emitted by CodeGen ($\mathcal{G}$), with $\mathtt{RC}$, $\mathtt{FV}$, $\mathtt{DA}$ abbreviating \texttt{rule\_checker}, \texttt{format\_validator}, \texttt{data\_analyzer}; $s_c = \mathtt{DA}(c.\mathtt{data})$ if $\mathtt{DA} \in \mathcal{M}_c$ and $\varnothing$ otherwise (the cached summary is additionally gated by an IR-strategy condition before injection into Reasoner-1; see §\ref{sec:reasoner-1-inputs}); $\tau(\cdot)$ is the head/tail context-truncation operator; $\mathcal{V}_c = \mathcal{M}_c \cap \{\mathtt{RC}, \mathtt{FV}\}$ is the draft-time verifier subset; and $d_t$ is the draft emitted by Reasoner-1 ($\mathcal{R}_1$). The second per-task factor $P(y_t | d_t, \mathcal{V}_c, \theta)$ bundles draft-time verifier execution and conditional correction by Reasoner-2 ($\mathcal{R}_2$): it computes the violation list $v_t \!=\! \bigcup_{m \in \mathcal{V}_c} m(d_t)$, sets $y_t \!=\! \mathcal{R}_2(d_t, v_t)$ when $|v_t| \!\geq\! \theta$ (we use $\theta \!=\! 2$) and $\mathcal{R}_2$ returns non-empty text, and otherwise sets $y_t \!=\! d_t$. By convention every $P(\cdot)$ implicitly conditions on the corresponding LLM stage's instruction prompt in $\mathcal{I}$; we write $\mathcal{I}$ explicitly only on the first term to declutter the~rest.

Equation~\ref{eq:cca} makes three structural properties of CCA explicit. \emph{(i)} The per-context artifacts $(I_c, \mathcal{M}_c, s_c)$ are amortized over every task that shares $c$; only the per-task product re-fires as $t$ varies. \emph{(ii)} The IR $I_c$ enters the Reasoner-1 prompt directly as a checklist (F2) and the cached pre-execution result $s_c$ enters as a separate input block (F5). \emph{(iii)} The draft-time verifiers $\mathcal{V}_c$ operate on $d_t$ (not on $c$); $\mathtt{DA}$ does not appear in the violation list because it has already been consumed in producing~$s_c$.

The labels F1--F7 attached to Eq.~\ref{eq:cca} are the unit of analysis in the ablation of §\ref{sec:rq1-answer}: F1 Compiler, F2 IR-as-Checklist Injection, F3 CodeGen Dispatcher, F4 Verifier Execution, F5 Data-Analyzer Pre-Pass, F6 Compile-then-Verify (correction loop), F7 Head/Tail Truncation. A per-F-code definition table with section back-references is in Appendix~\ref{app:feature-codes-list}.
\label{sec:feature-codes}

\subsection{Stage 1: Compiler}
\label{sec:compiler}

The Compiler reads the raw context (concatenated system + user messages) and emits a JSON IR whose fields (\texttt{role}, \texttt{rules}, \texttt{knowledge}, \texttt{workflow}, \texttt{output\_spec}, \texttt{available\_tools}, \texttt{data\_profile}, \texttt{compilation\_meta}) are summarised in Appendix~\ref{app:impl-ir-schema}. The Compiler prompt instructs the model to preserve fictional or counterfactual content verbatim (CL-bench contains contexts where domain facts are deliberately altered) and to extract every must/should/always/never statement as a rule. Two IR fields warrant emphasis: \texttt{knowledge.exact\_terms} captures verbatim strings (agent role names, status codes, identifiers) that downstream rubrics typically require literally; and the \texttt{codeable} flag on each rule decides whether it will be checked mechanically by CodeGen (§\ref{sec:codegen}) or left to the Reasoner's semantic judgment.

\subsection{Stage 2: CodeGen}
\label{sec:codegen}

The dispatcher inspects three IR fields (the count of codeable rules, \texttt{data\_profile.format}, and the number of \texttt{formatting\_rules} in \texttt{output\_spec}) and emits up to three Python modules. Two are draft-time verifiers that run \emph{post-Reasoner-1} on the draft response and emit a violation list (each tagged with rule\_id and evidence span): \texttt{rule\_checker(response\_text)}, emitted when the IR contains $\geq$$1$ rule with \texttt{codeable}=\texttt{True}; and \texttt{format\_validator(response\_text)}, emitted when the IR contains $\geq$$3$ \texttt{formatting\_rules}. The third, \texttt{data\_analyzer(raw\_data)}, is qualitatively different: it is emitted when \texttt{data\_profile.format} $\in$ \{\texttt{tsv}, \texttt{csv}, \texttt{inline\_table}\}, runs at the \emph{per-context} phase on the raw data block (not on any draft), and its cached summary is injected as a ``CODE EXECUTION RESULTS'' block in the Reasoner-1 prompt when the strategy is \texttt{data\_analysis} or \texttt{calculator}. Each module is generated via a separate LLM call whose prompt emphasises false-positive avoidance (flagging a non-violation would cause Correction to ``fix'' a correct draft); modules are wrapped in \texttt{try/except} so any code failure produces an empty result rather than crashing the pipeline.

\subsection{Stage 3: Reasoner-1}
\label{sec:reasoner-1-inputs}

The Reasoner-1 prompt concatenates, in order, four blocks: (1) a \textbf{system prompt} (the \texttt{REASONER\_SYSTEM} scaffold) that instructs the model to honor the context as the source of truth and the IR as a checklist of constraints; (2) a \textbf{compiled checklist} rendering of the IR (role persona, rules grouped by polarity, exact terms, workflow, output spec); (3) a conditional \textbf{code-execution-results} block, injected only when the IR's \texttt{compilation\_meta.recommended\_strategy} is \texttt{data\_analysis} or \texttt{calculator} and the per-context \texttt{data\_analyzer} module ran successfully (the draft-time verifiers \emph{never} run at this stage); and (4) the \textbf{original context}, truncated to the model's window via a head-and-tail policy ($\sim$$70\%$ head, $\sim$$30\%$ tail, with a padding marker), followed by the user's actual question. This separation lets the Reasoner focus on producing a correctly-formatted response that addresses the question while respecting the IR's constraints, without re-deriving the structural information already captured in the IR or repeating cached data-summarisation work.

\subsection{Stage 4: Verifier Execution and Reasoner-2 (Correction)}
\label{sec:reasoner-2}

After Reasoner-1 produces a draft, the \emph{Execute} step in Figure~\ref{fig:architecture} runs the draft-time verifier modules (\texttt{rule\_checker} and \texttt{format\_validator}, whichever were emitted) against the draft, collecting all violations into a single list (the per-context \texttt{data\_analyzer} module is not re-run here). If $n_{\text{viol}} \geq 2$ (chosen to suppress correction from a single noisy verifier hit), a Reasoner-2 \emph{Correction} call fires with three inputs: the violation list (truncated to the first 10 entries, 200 chars each), the original draft as a prior assistant message, and strict instructions to make only minimal local edits. The corrected output replaces the draft only when it is non-empty; otherwise the draft is retained, so Correction can only \emph{add} verified improvement, never silently break an otherwise-correct response.

Essential prompt cores for the four LLM-call stages and the IR schema spec are in Appendix~\ref{app:implementation}; the two-phase pipeline is also given as pseudocode in Appendix~\ref{app:algorithms} (Algorithms~\ref{alg:per-context}--\ref{alg:per-task}).

\section{Experiments and Results}
\label{sec:exp}

\subsection{Benchmark}

\textbf{CL-bench}~\citep{clbench} contains 1{,}899 tasks distributed across four domains and 18 sub-categories: Domain Knowledge Reasoning (DKR, 663 tasks), Rule System Application (RSA, 566), Procedural Task Execution (PTE, 471), and Empirical Discovery \& Simulation (EDS, 199). Each task consists of (i) a long context (median 20K characters, max 247K), (ii) a question, and (iii) a list of 5--20 rubric criteria. The rubric format follows ``must include / must not include / must be formatted as'', with each criterion graded independently and the task counted as pass only if every criterion is satisfied. We use the official rubric-based evaluation script that delegates per-criterion grading to a strong judge LLM.

\subsection{Base Models}

We evaluate four open LLMs accessed via AWS Bedrock: \textbf{Kimi K2.5}~\citep{moonshotai2026kimik25} (\texttt{moonshotai.kimi-k2.5}, 32B active / 1T total), \textbf{GLM-5}~\citep{zai2026glm5} (\texttt{zai.glm-5}, 40B active / 744B total), \textbf{DeepSeek-V3.2}~\citep{deepseekai2025deepseekv32} (\texttt{deepseek.v3.2}, 37B active / 671B total), and \textbf{Qwen3-Next-80B}~\citep{qwen2025qwen3next} (\texttt{qwen.qwen3-next-80b-a3b}, 3B active / 80B total). All four support input windows long enough to accommodate CL-bench's 247K-character maximum, so no input truncation is needed at the model boundary. All models receive identical prompts; temperature is fixed at 0.0 across all methods for reproducibility; output cap is 8{,}192 tokens (16{,}384 for DeepSeek-V3.2 per its native maximum).

\begin{table*}[t]
\centering\footnotesize
\setlength{\tabcolsep}{6pt}
\begin{tabular*}{\textwidth}{@{\extracolsep{\fill}}llcccccr@{}}
\toprule
\textbf{Method} & \textbf{Model} & \textbf{DKR} & \textbf{RSA} & \textbf{PTE} & \textbf{EDS} & \textbf{Overall} & \textbf{Tok/task} \\
\midrule
\multirow{4}{*}{Vanilla}
  & Kimi K2.5  & 16.7 & 13.4 & 17.2 & \textbf{12.1} & 15.4 & \textbf{11.9K} \\
  & GLM-5       & 17.9 & 15.2 & 17.2 & 10.1 & 16.1 & \textbf{12.0K} \\
  & DeepSeek   & 16.6 & 13.4 & 16.1 & \textbf{11.6} & 15.0 & \textbf{11.7K} \\
  & Qwen3      & 14.2 & 11.0 & 10.4 & \textbf{10.6} & 11.9 & \textbf{12.9K} \\
\midrule
\multirow{4}{*}{ReadAgent-P}
  & Kimi K2.5  & 14.3 & 15.4 & 17.8 &  6.5 & 14.7 & 14.0K \\
  & GLM-5       & 13.3 & 15.0 & 13.4 &  5.5 & 13.0 & 13.1K \\
  & DeepSeek   & 13.9 & 11.7 & 16.6 &  6.5 & 13.1 & 13.2K \\
  & Qwen3      & 12.8 & \textbf{11.1} & 12.7 &  2.5 & 11.2 & 15.2K \\
\midrule
\multirow{4}{*}{Ctx2Skill$^{\dag}$}
  & Kimi K2.5  & 19.3 & 13.3 & 15.1 & 10.1 & 15.5 & 193.9K \\
  & GLM-5       & 17.6 & 13.3 & 14.6 & \textbf{14.1} & 15.2 & 191.4K \\
  & DeepSeek   & 16.1 & 13.4 & 16.8 & \textbf{11.6} & 15.0 & 183.0K \\
  & Qwen3      & 14.0 &  9.9 & 11.9 & \textbf{10.6} & 11.9 & 193.3K \\
\midrule
\multirow{4}{*}{\textbf{CCA (ours)}}
  & Kimi K2.5  & \textbf{21.1} & \textbf{21.2} & \textbf{26.1} & 11.6 & \textbf{21.4} & 34.0K \\
  & GLM-5       & \textbf{23.5} & \textbf{19.8} & \textbf{25.1} &  8.0 & \textbf{21.2} & 30.6K \\
  & DeepSeek   & \textbf{17.8} & \textbf{16.8} & \textbf{22.5} &  9.0 & \textbf{17.7} & 29.3K \\
  & Qwen3      & \textbf{15.4} &  9.7 & \textbf{14.6} &  4.5 & \textbf{12.4} & 34.1K \\
\bottomrule
\end{tabular*}
\caption{Per-domain pass rate (\%) and average tokens per task on CL-bench. Domain codes DKR / RSA / PTE / EDS are defined in §\ref{sec:exp}. \textbf{Bold} pass rates flag the best score per (model, domain) cell (ties both bolded); bold tokens flag the cheapest method per row. Tok/task sums all LLM-call tokens (input $+$ output) per task, with any per-context offline stage amortised over tasks-per-context (Ctx2Skill amortises its self-play stage across all 1{,}899 tasks; ${}^{\dag}$its inference-only cost is $\sim$$15$K/task, comparable to Vanilla). CCA achieves the best (model, Overall) cell on every base model while spending $\sim$$5.7\times$ fewer tokens per task than Ctx2Skill, the other offline-prep$+$online-infer method in the suite; full cost decomposition in Appendix~\ref{app:cost}.}
\label{tab:main-results}
\end{table*}

\subsection{Baselines}

We compare CCA against three baselines spanning the major ICL strategies (full porting details and adaptation notes in Appendix~\ref{app:baselines}):

\paragraph{Vanilla.} Direct prompting: the full original context is sent to the base model and the response is returned with no preprocessing or post-processing.

\paragraph{ReadAgent-P} \citep{lee2024readagent}. A long-context retrieval method using episode pagination, page-level gisting, and parallel look-up. We port the official Colab implementation, using its prompts verbatim (only the answer prompt is adapted from QuALITY's multiple-choice format to CL-bench's free-form format). Pagination and gisting are amortized per context.

\paragraph{Ctx2Skill} \citep{si2026ctx2skill}. A multi-agent self-play method using Challenger, Reasoner, Judge, Proposer, and Generator agents to evolve a skill library across iterations. We port the official codebase to Bedrock and run with the default configuration: 5 iterations $\times$ 5 tasks per iteration $\times$ 500 unique contexts, sharing the same four base~models.

\subsection{Evaluation Protocol}

We evaluate all graded outputs using GPT-5.1 (\texttt{gpt-5.1}) as the judge with the official CL-bench script, which returns a per-criterion yes/no plus an overall $0$/$1$ score. Inference temperature is fixed at $0.0$ across all methods for reproducibility and clean ablation attribution, removing sampling noise from per-component deltas so that lifts can be attributed to design choices rather than run-to-run variance; the judge uses default sampling. As a side effect our absolute numbers sit modestly below the $23.7\%$ best reported in the original CL-bench paper~\citep{clbench}, which evaluates a partially different (larger, closed-model-inclusive) model pool under default sampling; CCA's lifts over Vanilla under our stricter temp$=0$ regime are correspondingly conservative. To prevent truncation bias we ran a two-pass audit (176 close-to-cap suspects re-run at each model's native maximum; post-audit no output lies within $10\%$ of its raised ceiling; full details in Appendix~\ref{app:audit}). To verify CCA's lift is not a greedy-decoding artifact, re-running Full CCA on Kimi K2.5 at temperature $1.0$ drops pass rate modestly to $18.75\%$ ($-2.65$pp vs temp$=0$) while still beating Vanilla at temp$=0$ by $+3.35$pp (full breakdown in Appendix~\ref{app:temperature-robustness}).

\subsection{Results}
\label{sec:results}

We organize the findings against the three research questions from §\ref{sec:intro}. Table~\ref{tab:main-results} reports pass rates for every (method, model, domain) cell plus the Overall pass rate over all 1{,}899 tasks.

\paragraph{RQ1 --- effectiveness of context compilation (first pass).} CCA achieves the highest Overall pass rate on every base model. On the three larger-activation models, CCA's lift over Vanilla is substantial: \textbf{+6.0pp on Kimi K2.5} (15.4$\rightarrow$21.4\%), \textbf{+5.1pp on GLM-5} (16.1$\rightarrow$21.2\%), and \textbf{+2.7pp on DeepSeek-V3.2} (15.0$\rightarrow$17.7\%). All three improvements are highly significant by paired McNemar test ($z = 6.00,\,4.89,\,3.01$; $p < 0.01$; test statistics in Appendix~\ref{app:mcnemar}). On the smallest model (Qwen3-Next-80B, 3B active) the lift drops to +0.5pp and is not significant ($z = 0.38$, $p = 0.70$); we revisit this model-capacity boundary as part of RQ3 in §\ref{sec:discussion}.

\paragraph{RQ2 --- comparison with existing strategies (first pass).} In every Overall column of Table~\ref{tab:main-results}, neither ReadAgent-P (gist-memory retrieval) nor Ctx2Skill (multi-agent self-play skill library) reaches CCA's score on any base model; the Tok/task column further shows that Ctx2Skill's similar offline-prep+online-infer architecture costs $\sim$$5.7\times$ more tokens per task than CCA, visualised as a Pareto-dominance pattern in Figure~\ref{fig:cost-quality}. We unpack the design-property mechanism in §\ref{sec:rq2-answer} (porting details in Appendix~\ref{app:baselines}; cost decomposition in §\ref{sec:cost} and Appendix~\ref{app:cost}).

\begin{figure}[t]
\centering
\includegraphics[width=\columnwidth]{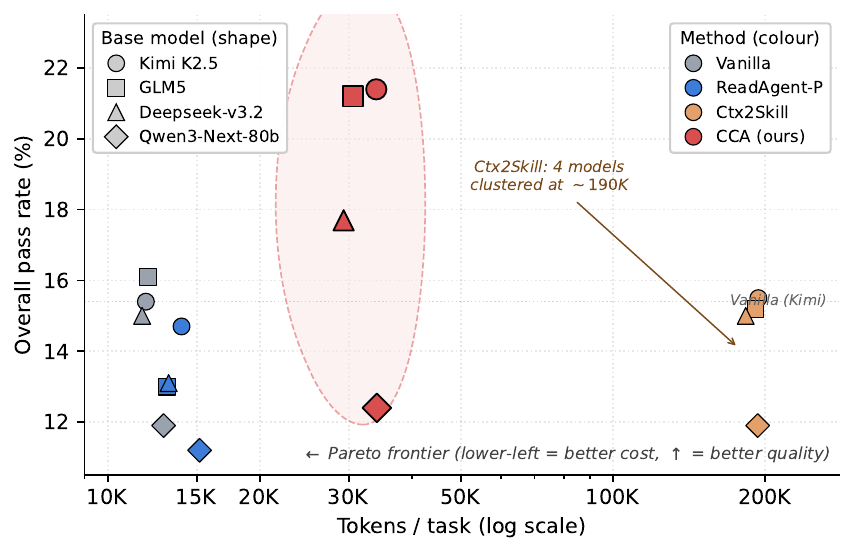}
\caption{Cost--quality tradeoff for the $16$ (method, model) cells of Table~\ref{tab:main-results}. Tokens per task on log $x$; Overall pass rate on $y$. \textbf{CCA (red) dominates the upper-middle Pareto frontier}: at ${\sim}3\times$ Vanilla's tokens it adds $+6.0/+5.1/+2.7/+0.5$pp on Kimi/GLM-5/DeepSeek/Qwen3 (last is capacity-bounded; see §\ref{sec:rq3-answer}). Ctx2Skill is the counter-example---${\sim}5.7\times$ CCA's cost, near-Vanilla pass rate---the lossy-abstraction failure of §\ref{sec:rq2-answer}. A cost-conscious CCA variant (no R-2; $+4.72$pp at ${\sim}22.7$K tokens) sits further left; see Appendix~\ref{app:cost}.}
\label{fig:cost-quality}
\end{figure}

\paragraph{RQ3 --- where the lift holds (first pass).} CCA's lift concentrates on rule-dense domains (PTE, RSA) where compiled rules and exact-term extraction directly support the rubric structure (CCA wins all four PTE columns and three of four RSA columns); DKR lift is smaller but still positive on every model. On open-ended EDS, CCA does not win on any model. The 18-sub-category lift table is in Appendix~\ref{app:full-matrix}; four CCA-only-wins case studies (one per domain) are in Appendix~\ref{app:case-studies}; we synthesise the domain-density and model-capacity moderators in §\ref{sec:rq3-answer}.

\paragraph{Cross-benchmark probe (LongBench-v2).} As a cross-benchmark generalization probe, we ran the 4-way comparison on LongBench-v2 (503 multi-choice tasks; matched \texttt{max\_tokens}=16{,}384 across all methods). Aggregate: CCA 53.88\% vs Vanilla 57.85\%; domain-selective wins on Multi-Doc QA (+5.69pp) and Long-Dialog (+10.25pp) --- see \S\ref{sec:when-to-use} for the full breakdown.

\subsection{Cost}
\label{sec:cost}

CCA splits into per-context \emph{offline prep} (Compiler+CodeGen+\texttt{data\_analyzer}) and per-task \emph{online infer} (Reasoner-1; Reasoner-2 on $\geq\!2$ violations). On Kimi K2.5, offline is $6.4$K tokens amortised over $\bar N\!=\!5.13$ tasks/ctx; online is $27.7$K per task. Verifier code is deterministic local Python at zero LLM tokens. Wall-clock is $\sim$$1.6\times$ Vanilla (below the $\sim$$2.85\times$ token ratio: Compiler/CodeGen fire once per group; CodeGen modules parallelise). \textbf{CCA-V2} (verifier check, no correction) gives $+4.72$pp at $1.90\times$ tokens. Table~\ref{tab:cost-per-n} shows the per-$N$ amortisation dial (extended in Appendix~\ref{app:cost}).

\begin{table}[t]
\centering
{\small
\begin{tabular}{@{}lrrr@{}}\toprule
$N$ & Tok/task & Calls & Cost$\times$ \\ \midrule
$1$ & $51.1$K & $5.03$ & $4.28\times$ \\
$5$ & $32.3$K & $2.29$ & $2.71\times$ \\
$10$ & $30.0$K & $2.05$ & $2.51\times$ \\
$\infty$ & $27.7$K & $1.60$ & $2.32\times$ \\ \bottomrule
\end{tabular}}
\caption{Per-context amortisation of CCA cost as a function of tasks-per-context $N$ (Kimi K2.5).}
\label{tab:cost-per-n}
\end{table}

\subsection{Reproducibility}
\label{sec:repro}
We release cached completions (4 methods$\times$4 models$\times$1{,}899 CL-bench + 503 LongBench-v2), runners/adapters, CCA-Adaptive gate ($\sim$30 lines), per-item failure-mode labels, and multi-judge (Grok/Gemma/Mistral) grades with $\kappa$; zero-Bedrock re-grading. All code and data are available at \url{https://github.com/TonyQJH/cca-emnlp2026}.

\section{Discussion}
\label{sec:discussion}

This section synthesizes our findings against each of the three research questions from §\ref{sec:intro}.

\subsection{Answering RQ1: Effectiveness of Context Compilation}
\label{sec:rq1-answer}

\textbf{RQ1 asked whether explicitly compiling context into a structured IR with executable verifiers improves rubric-graded ICL pass rate.} The main-table results in §\ref{sec:results} already answer affirmatively for every base model (significant by McNemar test on Kimi K2.5, GLM-5, and DeepSeek-V3.2). The deeper question is \emph{which component} of CCA delivers the lift; we localise it via a controlled ablation on Kimi K2.5 against the full 1{,}899-task set, disabling one feature (or feature group) at a time from Full CCA (Table~\ref{tab:ablation}).

\begin{table*}[t]
\centering\small
\setlength{\tabcolsep}{6pt}
\renewcommand{\arraystretch}{1.18}
\begin{tabular*}{\textwidth}{@{}l@{\extracolsep{\fill}}cccc@{}}
\toprule
\textbf{Configuration} & \textbf{Pass} & \textbf{Acc.} & \textbf{$\Delta$ vs Full} & \textbf{$\Delta$ vs Vanilla} \\
\midrule
\textbf{Full CCA} (all features on)                                & \textbf{406}/1899 & \textbf{21.40\%} & ---       & $\mathbf{+6.00}$ \\
\midrule
\multicolumn{5}{@{}l}{\textit{Single-feature ablations}} \\
\quad CCA w/o F6 (Reasoner-2 correction loop)                       & 382/1899 & 20.12\% & $-1.28$   & $+4.72$ \\
\quad CCA w/o F2 (IR-as-checklist injection)$^{\star}$              & 348/1899 & 18.33\% & $\mathbf{-3.07}$ & $+2.93$ \\
\midrule
\multicolumn{5}{@{}l}{\textit{Multi-feature ablations}} \\
\quad CCA w/o F4 + F6 (verifier execution + correction)             & 372/1899 & 19.59\% & $-1.81$   & $+4.19$ \\
\quad CCA w/o F1 cascade (Compiler off $\to$ F2/F4/F5/F6 disabled)  & 327/1899 & 17.22\% & $-4.18$   & $+1.82$ \\
\midrule
Vanilla (no CCA pipeline; default prompting)                        & 292/1899 & 15.40\% & $-6.00$   & --- \\
\bottomrule
\end{tabular*}
\caption{Ablation on Kimi K2.5 over the full 1{,}899 CL-bench tasks. Each row removes one CCA feature (or group) from Full CCA, keeping the rest active. ${}^{\star}$Removing F2 (IR-as-checklist injection) causes the \emph{single largest drop} ($-3.07$pp), identifying F2 as the most load-bearing feature. F-codes are defined in Appendix~\ref{app:feature-codes-list}; the four drops recombined as an additive decomposition (§\ref{sec:rq1-answer}) sum exactly to the $+6.00$pp Full-vs-Vanilla gap.}
\label{tab:ablation}
\end{table*}

\paragraph{F2 is the load-bearing component.} Reading the single-feature ablations in Table~\ref{tab:ablation}, \textbf{removing F2 (IR-as-checklist injection) causes the largest single drop} ($21.40 \to 18.33\%$, $-3.07$pp), nearly $2.5\times$ the drop from removing F6 (Reasoner-2 correction; $-1.28$pp). Surfacing the typed IR (rules grouped by polarity, verbatim exact-terms, output schema) inside the Reasoner-1 prompt is what most heavily lifts pass rate, even before any verifier code runs. F4+F6 jointly drop $-1.81$pp, so F4 alone (without correction) contributes the small remainder $\sim$$0.53$pp: violations are logged but not acted on. Disabling the Compiler cascades F2/F4/F5/F6 off, dropping $-4.18$pp; the residual $+1.82$pp Vanilla-to-V6 gap measures the prompt scaffold plus F7 truncation alone. Re-combined in build-up order, these contributions form an additive decomposition that sums exactly to the $+6.00$pp gap: $+1.82$ (scaffold) $+\,2.37$ (F2 IR-as-Checklist Injection) $+\,0.53$ (F4 Verifier Execution) $+\,1.28$ (F6 Compile-then-Verify); Figure~\ref{fig:ablation_buildup} visualises this with the F2 step highlighted as the largest jump.

\paragraph{Where CCA's value lives.} The IR-as-checklist injection is the dominant single mechanism; the correction loop is second; the verifier execution alone (without correction) is essentially a logging step; the Compiler is the enabling foundation, realised through F2/F4/F6 rather than as a standalone benefit. F2 also interacts positively with the F4/F6 loop ($+0.70$pp interaction term; derivation in Appendix~\ref{app:extended-ablation}), so F2's measured contribution is larger when removed from Full CCA ($-3.07$pp) than when added on top of scaffold-only ($+2.37$pp). Per-domain ablation, fire rates, token costs, and the build-up visualisation (Figure~\ref{fig:ablation_buildup}) are in Appendix~\ref{app:extended-ablation}; the cost-quality dial is in Appendix~\ref{app:cost}.

\subsection{Answering RQ2: Why Existing Strategies Don't Improve}
\label{sec:rq2-answer}

\textbf{RQ2 asked how context compilation compares to existing long-context strategies and which design properties of each are well- or poorly-matched to rubric-based grading.} The shared failure mode of the baselines is a \emph{lossy abstraction step} between context and reasoner: ReadAgent-P's gisting summarises away verbatim facts; Ctx2Skill's skill distillation summarises away per-criterion checks. Rubric grading penalises both losses (it requires verbatim reproduction of specific terms and binary satisfaction of each criterion), so neither baseline can match a method that keeps those signals intact. CCA's design is the inverse: the IR preserves verbatim \texttt{exact\_terms} and CodeGen produces per-context executable checks; both are tightly coupled to what the rubric tests, which is why CCA beats both baselines on every (model, Overall) cell of Table~\ref{tab:main-results} and dominates the Pareto frontier of Figure~\ref{fig:cost-quality}: the off-frontier baselines pay extra tokens over Vanilla but return near-zero lift, while CCA converts its $\sim$$3\times$ Vanilla cost into the only large positive lift (full positioning vs all three baselines in Appendix~\ref{app:cost}). Appendix~\ref{app:case-studies} traces four CCA-only-wins (one per domain) end-to-end, showing exactly which IR field or verifier module differentiates CCA from each baseline on the same task.

\subsection{Answering RQ3: Where the Compilation Harness Holds}
\label{sec:rq3-answer}

\textbf{RQ3 asked where the harness benefit holds, and how it depends on the underlying model's capacity and the task's structural density.} Two moderators emerge.

\paragraph{Moderator 1 (task structural density).} The largest sub-category lifts cluster where the context defines an explicit rule system (full 18-sub-category lift table in Appendix~\ref{app:subcat-lifts}): \textbf{Legal \& Regulatory} (RSA, $n\!=\!92$) gains $+20$ to $+27$pp across all models (Compiler extracts statute lists and jurisdictions as \texttt{exact\_terms}); \textbf{Management} (DKR, $n\!=\!112$) gains $+11$ to $+25$pp (the IR's \texttt{rules} and \texttt{output\_spec.persona\_voice} together encode the management persona's decision framework); \textbf{Workflow Orchestration} (PTE, $n\!=\!229$) gains $+6$ to $+11$pp across all four models (rich \texttt{available\_tools} and \texttt{workflow} steps are captured explicitly). Conversely, CCA underperforms Vanilla on open-ended creative tasks (e.g.\ Game Mechanics, Lifestyle, Simulation Environment), where the compiled IR acts as a Procrustean bed: the Reasoner produces a structurally clean but content-narrow response while Vanilla rambles freely and happens to cover more of what the grader expects. We view this not as a failure of the compilation principle but as a need to vary IR strictness as a function of \texttt{compilation\_meta}'s recommended~strategy.

\paragraph{Moderator 2 (model capacity).} The lift on Qwen3-Next-80B (3B active) drops to $+0.5$pp and is not significant ($p\!=\!0.70$), suggesting that harness benefit is bounded by the underlying model's capacity to integrate structured supplementary signal: a small-activation model cannot fully consume an IR checklist plus injected code-execution results \emph{in addition to} the original context. On Kimi K2.5 the Reasoner-2 correction fires on $\sim$$62\%$ of tasks (Appendix~\ref{app:extended-ablation}, Table~\ref{tab:appendix-correction-fire}); on the small-activation model the corrected output is empty more often (12 empty-output cases concentrated on Qwen3 in our audit, Appendix~\ref{app:audit}), so the draft is retained; the conservative ``only add verified improvement'' design effectively no-ops on Qwen3. Model-capacity scaling laws are left to future~work~(§\ref{sec:limitations}).

\subsection{When to Use CCA: A Domain-Selective Harness}
\label{sec:when-to-use}
CCA is a domain-selective tool, not a universal wrapper. On LongBench-v2 (Kimi K2.5, $n\!=\!503$, matched \texttt{max\_tokens}$=$16{,}384 across all methods), pure CCA trails Vanilla in aggregate but wins on the two multi-context reconciliation domains that the typed IR was architecturally designed to help:

\begin{center}
{\small
\begin{tabular}{@{}lccc@{}}\toprule
Method & Multi-Doc & Long-Dialog & Overall \\
       & QA ($n\!=\!123$) & ($n\!=\!39$) & ($N\!=\!503$) \\ \midrule
Vanilla        & 52.85 & 53.85 & \textbf{57.85} \\
Ctx2Skill      & 52.03 & 58.97 & 55.86 \\
ReadAgent-P    & 53.66 & 58.97 & 54.08 \\
Pure CCA       & \textbf{58.54} & \textbf{64.10} & 53.88 \\ \midrule
$\Delta$ vs Vanilla & $+5.69$ & $+10.25$ & $-3.97$ \\ \bottomrule
\end{tabular}}
\end{center}

The pattern is the paper's mechanism prediction landing on an unrelated benchmark: typed IR helps where structure must be assembled across sources or turns, and adds overhead on single-context tasks where ``just read it'' suffices. \textbf{CCA-Adaptive} operationalises this asymmetry as a one-line runtime gate on \texttt{cca\_meta.ir\_ok} (apply CCA when the Compiler emits verifier modules, fall back to Vanilla otherwise), lifting the aggregate to \textbf{60.24\%} ($+2.39$pp vs Vanilla).

\section{Conclusion}
\label{sec:conclusion}

We introduce \textbf{CCA}, whose central contribution is a typed intermediate representation (IR) with fixed slots that turns unstructured prose context into a first-class, machine-checkable data structure; verifier synthesis and violation-gated correction follow as downstream consequences. On CL-bench, CCA attains the highest pass rate on every base model: $+6.0/+5.1/+2.7$pp over Vanilla on Kimi K2.5 / GLM-5 / DeepSeek-V3.2 (all $p<0.01$), and a smaller, non-significant $+0.5$pp on Qwen3-Next-80B, the capacity-bounded case (§\ref{sec:rq3-answer}); CCA also outperforms ReadAgent-P and Ctx2Skill. On LongBench-v2, \textbf{CCA-Adaptive} (a \texttt{cca\_meta.ir\_ok}-gated router) reaches $60.24\%$, $+2.39$pp over Vanilla (§\ref{sec:when-to-use}). The harness-engineering framing positions CCA as an auto-derived production harness around a frozen base LLM, supporting the broader claim that progress on long-context rubric-graded benchmarks benefits more from compiling structure \emph{out of} the context than from training models to better self-recall.

\section{Limitations}
\label{sec:limitations}

\textbf{Scaling the ablation.} Our component-level isolation (§\ref{sec:rq1-answer}) is run on Kimi K2.5 over the full 1{,}899-task set. Running the same isolation across the other three base models (particularly probing the F2/F6 interaction at smaller activation, §\ref{sec:rq3-answer}) would sharpen the model-capacity boundary at which harness benefit saturates, and is a natural extension of this work.

\textbf{Schema coverage.} The current IR captures the structural primitives that recur across rule-heavy, procedural, and data-analysis contexts: rules, exact terms, output specs, agent workflows, and tabular data profiles. We see direct extensions to contexts whose relevant structure is \emph{spatial} (geometric reasoning, layout) or \emph{temporal} (long-horizon planning, event ordering) .

\textbf{Compiler robustness.} Unparseable IRs occurred for $<1\%$ of context groups in our evaluation, all absorbed by graceful fallback to direct prompting; hardening this into a formally verified property for safety-critical adversarial deployments is left to future work.

\paragraph{Domain-selective on open-ended long-context.} On LongBench-v2 fair 16K, CCA's aggregate accuracy (53.88\%) trails Vanilla (57.85\%); wins are concentrated on multi-context reconciliation domains (Multi-Doc QA +5.69pp, Long-Dialog +10.25pp) and vanish on single-context tasks. CCA-Adaptive (\S\ref{sec:when-to-use}) partially addresses this via a one-line runtime gate on \texttt{cca\_meta.ir\_ok}, but the underlying limitation is that typed IR extraction adds overhead where structure cannot be extracted.

\paragraph{Single-judge scoring.} All CL-bench pass rates are scored by a single GPT-5.1 judge following the rubric protocol of \citet{clbench} (\S3.4 there). Cross-judge triangulation (Grok 4.3, Gemma 4 31B, Mistral Large 3) on a stratified 500-task subset shows GPT-5.1 is the strictest of four frontier judges (pairwise $\kappa$ 0.40--0.61), so the paper's absolute pass rates are conservative measurements; a full 100-task human audit is left to future work.

\section*{Acknowledgements}
The research presented in this paper was partially supported by the Research Grants Council of the Hong Kong Special Administrative Region, China (CUHK 2300246, RGC C1043-24G), (CUHK 14203425, RGC GRF 2151317), and CUHK 7010886.

\bibliography{custom}

\appendix
\section{Implementation Reference: Prompt Cores and IR Schema}
\label{app:implementation}

\begingroup
\lstset{basicstyle=\ttfamily\footnotesize,frame=single,backgroundcolor=\color{gray!10},breaklines=true,xleftmargin=6pt,framexleftmargin=4pt,framexrightmargin=4pt,aboveskip=6pt,belowskip=4pt}

This appendix documents the essential portions of each LLM-call prompt used in CCA, plus the IR schema produced by the Compiler. Boilerplate (formatting reminders, "no markdown fences", general politeness) is elided; the goal here is to expose the \emph{design-relevant} instructions that, if changed, would change pipeline behaviour. Full prompts and code are released with the supplementary material.

\subsection{Compiler Prompt --- Essential Sections}
\label{app:impl-compiler}

The Compiler is the load-bearing per-context stage. Three sections of its prompt are essential for the pipeline's behaviour:

\paragraph{(1) Context fidelity (counterfactual preservation).}\mbox{}\\[-0.5em]
\begin{lstlisting}
Context may contain fictional/modified/
counterfactual knowledge. Extract EXACTLY
as stated, even when contradicting your
training. Do not "correct" or skip rules
that seem unusual.
\end{lstlisting}
This instruction is what lets CCA pass CL-bench tasks whose contexts deliberately use counterfactual rules.

\paragraph{(2) \texttt{exact\_terms} extraction.}\mbox{}\\[-0.5em]
\begin{lstlisting}
Rubrics frequently check for VERBATIM
appearance of specific strings from
the context. Extract every term that
downstream output must reproduce
literally:
- Agent role identifiers
- Tool/command names
- Status codes / enum values
- Specific identifiers (e.g. MEDIA-002)
- Rating tier names
- Exact phrases the persona must use

DO NOT paraphrase or generalize ---
the rubric is checking for the
literal string.
\end{lstlisting}
This drives the dominant ablation finding: F2 (IR-as-checklist injection) is the largest single contributor, primarily because the \texttt{exact\_terms} field surfaces strings the rubric will literally check for.

\paragraph{(3) \texttt{output\_spec.format\_type} classification.} The Compiler tags each context with one of \{\texttt{natural\_artifact}, \texttt{tool\_sequence}, \texttt{structured\_data}, \texttt{hybrid}\} (see Table~\ref{tab:ir-schema}), used by Reasoner-1 to choose response format. The classifier prompt enumerates examples for each label and defaults to \texttt{natural\_artifact} on uncertainty.

\subsection{IR Schema}
\label{app:impl-ir-schema}

Table~\ref{tab:ir-schema} summarises every top-level field of the JSON IR emitted by the Compiler. Each rule under \texttt{rules.must\_do}/\texttt{must\_not}/\texttt{conditional} carries an \texttt{id}, the verbatim \texttt{rule} text, the \texttt{source} quote it was extracted from, a boolean \texttt{codeable} flag (whether CodeGen should attempt to verify it mechanically), and an optional \texttt{code\_hint} string (a regex or check the CodeGen prompt is encouraged to reuse). Each item in \texttt{knowledge.exact\_terms} carries a \texttt{category}, the verbatim \texttt{term}, and an \texttt{example\_use} describing how the rubric is expected to test for it.

\begin{table}[t]
\centering\small
\setlength{\tabcolsep}{4pt}
\begin{tabular}{@{}lp{0.58\linewidth}@{}}
\toprule
\textbf{IR field} & \textbf{Contents} \\
\midrule
\texttt{role}             & persona, voice, tone \\
\texttt{rules}            & \texttt{must\_do}/\texttt{must\_not}/\texttt{conditional}; each with a \texttt{codeable} flag and \texttt{code\_hint} \\
\texttt{knowledge}        & concepts, entities, formulas, \texttt{exact\_terms} (verbatim strings rubrics require) \\
\texttt{workflow}         & ordered procedural steps and decision points \\
\texttt{output\_spec}     & \texttt{format\_type} $\in$ \{\texttt{natural\_artifact}, \texttt{tool\_sequence}, \texttt{structured\_data}, \texttt{hybrid}\}, tone, \texttt{persona\_voice}, \texttt{formatting\_rules}, length \\
\texttt{available\_tools} & tool name, purpose, I/O schema, when-to-call \\
\texttt{data\_profile}    & \texttt{none / tsv / csv / inline\_table} + optional analysis hints (consumed by CodeGen) \\
\texttt{compilation\_meta} & recommended strategy $\in$ \{\texttt{rule\_engine}, \texttt{calculator}, \texttt{data\_analysis}, \texttt{soft\_reasoning}\} (consumed by Reasoner) \\
\bottomrule
\end{tabular}
\caption{IR fields emitted by the Compiler.}
\label{tab:ir-schema}
\end{table}

\subsection{CodeGen Prompt Cores}
\label{app:impl-codegen}

CodeGen emits up to three Python modules. The single design property they share is \emph{false-positive avoidance}:
\begin{lstlisting}
NO FALSE POSITIVES. Only flag a
violation if you are CERTAIN it's
violated. When in doubt, skip the
rule. False positives are catastrophic
--- they cause the downstream Reasoner
to "fix" things that aren't broken.
\end{lstlisting}

\paragraph{\texttt{rule\_checker}.} Generates \texttt{check\_rules(response\_text) -> dict} returning a violation list. The prompt enumerates common false-positive patterns to avoid: negated assertions, phrases inside quoted text, case sensitivity (default insensitive), word boundary leakage (e.g., matching ``art'' inside ``start'').

\paragraph{\texttt{format\_validator}.} Generates \texttt{check\_format(response\_text) -> dict} for formatting rules: ``must end with X'' $\to$ regex anchored at end-of-string; ``must include JSON field Y'' $\to$ \texttt{json.loads} + key check; ``must be \texttt{<} 200 words'' $\to$ word-count split.

\paragraph{\texttt{data\_analyzer}.} Generates \texttt{analyze\_data(raw\_data) -> dict}. Unlike the previous two, this module is executed at the per-context phase on the raw data block. Its core design property is \emph{defensive parsing} (real-world data is messy):
\begin{lstlisting}
DEFENSIVE PARSING --- try multiple
strategies in order: strip
BOM/whitespace; auto-detect delimiter
(tab, comma, semicolon, space, pipe);
pandas with sep=None,
engine='python', on_bad_lines='skip';
manual csv module on failure; per-row
column-count validation.
\end{lstlisting}
The module is wrapped in a top-level \texttt{try/except} that always returns partial results (never raises) so a malformed data block cannot poison the downstream pipeline.

\subsection{Reasoner-1 System Prompt --- Essential Sections}
\label{app:impl-reasoner1}

The system prompt is the constant component of every Reasoner-1 call. Two sections do the heavy lifting:

\paragraph{(1) Context-first principle (counterfactual handling).}\mbox{}\\[-0.5em]
\begin{lstlisting}
The CHECKLIST and ORIGINAL CONTEXT are
the AUTHORITATIVE source. When context
contradicts your prior knowledge ---
TRUST THE CONTEXT.
- Do NOT add disclaimers
  ("in reality...", "actually...",
   "note that the standard X is...")
- Do NOT correct, qualify, or
  fact-check the context using your
  training knowledge
- Apply the rules AS GIVEN, even if
  they seem strange.
\end{lstlisting}
This mirrors the Compiler's context-fidelity instruction at the Reasoner level: the structured IR alone is not enough --- the Reasoner must \emph{accept} the counterfactual content rather than ``fix'' it.

\paragraph{(2) Rule precedence within context.}\mbox{}\\[-0.5em]
\begin{lstlisting}
A specific situational rule beats a
general one. An explicit example in
context beats a generic principle
from your training. The rule that
mentions the EXACT entities/conditions
in the task wins.
\end{lstlisting}
This is the conflict-resolution layer for contexts containing overlapping rules, common in legal/regulatory and management domains.

\subsection{Reasoner-2 Correction Prompt}
\label{app:impl-reasoner2}

Correction is invoked as a follow-up turn with the original Reasoner-1 messages still in the conversation, plus the Reasoner-1 draft appended as a prior \texttt{assistant} turn, plus a new \texttt{user} message containing the violation feedback:
\begin{lstlisting}
## CODEABLE/FORMAT VIOLATIONS (n):
- [R3] {rule text, <=200 chars}
    Evidence: {snippet, <=200 chars}
- [N1] {rule text, <=200 chars}
    Evidence: ...
...

## YOUR DRAFT:
{Reasoner-1 draft, <=6000 chars}

These are objective, code-verifiable
violations. Revise your response to
fix them precisely. Do NOT make other
changes --- keep the structure, tone,
and content otherwise identical.
Output the revised response only.
\end{lstlisting}
The violation list is truncated to the first 10 violations (each <=200 characters for both rule text and evidence span) and the draft is truncated to <=6000 characters, to stay within the model's prompt budget. The instruction to make ``only the flagged edits'' is essential: it is what makes correction monotone --- the corrected output strictly fixes at least one verified violation, rather than reorganising the draft in ways that might introduce new ones.

\subsection{Head/Tail Context Truncation (F7)}
\label{app:impl-truncation}

When the original context block exceeds the Reasoner's input budget (set to 200{,}000 characters in our runs), we truncate it as follows: keep the first $70\%$ of the budget as ``head'' (preserves the system prompt and rule definitions, which are typically front-loaded in CL-bench contexts) and the last $30\%$ as ``tail'' (preserves the user's actual question and any most-recent turn), with a 200-character padding marker \texttt{"[\ldots context truncated \ldots]"} inserted between them. This policy is the F7 component of the ablation (§\ref{sec:rq1-answer}); its contribution on Kimi K2.5 is folded into the prompt-scaffold step ($+1.82$pp Vanilla $\to$ V6).

\subsection{F-Code Definitions for the Ablation}
\label{app:feature-codes-list}

The labels F1--F7 attached to Eq.~\ref{eq:cca} and used throughout §\ref{sec:rq1-answer} are defined here with cross-references to their detailed treatment in the main text.
\begin{itemize}
\setlength{\itemsep}{0pt}
\item \textbf{F1 --- Compiler} (§\ref{sec:compiler}): per-context LLM call that emits the typed JSON IR $I_c$.
\item \textbf{F2 --- IR-as-checklist injection} (§\ref{sec:reasoner-1-inputs}): renders $I_c$ (rules, exact terms, workflow, output spec) into the Reasoner-1 prompt.
\item \textbf{F3 --- CodeGen dispatcher} (§\ref{sec:codegen}): per-context LLM call(s) deciding the module set $\mathcal{M}_c$ and generating each module's Python source.
\item \textbf{F4 --- Draft-time verifier execution} (§\ref{sec:reasoner-2}, the ``Execute'' step): applies the modules in $\mathcal{V}_c$ to the Reasoner-1 draft and produces the violation list $v_t$.
\item \textbf{F5 --- Per-context $\mathtt{data\_analyzer}$} (§\ref{sec:codegen}): runs once on the raw context's data block in the per-context phase; its cached summary $s_c$ is injected into Reasoner-1's prompt as a separate block when the IR's recommended strategy is \texttt{data\_analysis} or \texttt{calculator}.
\item \textbf{F6 --- Reasoner-2 correction loop} (§\ref{sec:reasoner-2}): regenerates the response when $|v_t| \geq \theta$, with $v_t$ and $d_t$ as input.
\item \textbf{F7 --- Head/tail context truncation policy} (§\ref{sec:reasoner-1-inputs}): $\tau(\cdot)$ keeps $\sim$$70\%$ head and $\sim$$30\%$ tail of the original-context block in the Reasoner-1 prompt.
\end{itemize}
\endgroup

\section{Algorithm Pseudocode}
\label{app:algorithms}

We present pseudocode for the two phases of CCA introduced in §\ref{sec:pipeline-formalism} and visualised in Figure~\ref{fig:architecture}. The notation follows Equation~\ref{eq:cca}: $\mathcal{C}, \mathcal{G}, \mathcal{R}_1, \mathcal{R}_2$ are LLM-call operators; $\mathcal{I}$ is the bag of instruction prompts; $\mathtt{RC}, \mathtt{FV}, \mathtt{DA}$ are the three Python verifier modules; $\tau(\cdot)$ truncates the original context to fit the Reasoner-1 window; and $\theta=2$ is the violation threshold.

\begin{algorithm}[t]
\small
\caption{CCA Per-Context Phase}
\label{alg:per-context}
\begin{algorithmic}[1]
\REQUIRE raw context $c$; instruction prompts $\mathcal{I}$
\ENSURE per-context artifacts $(I_c, \mathcal{M}_c, s_c)$
\STATE $I_c \gets \mathcal{C}(c \mid \mathcal{I})$ \hfill\COMMENT{\textbf{F1}: Compiler emits JSON IR}
\STATE $\mathcal{M}_c \gets \emptyset$ \hfill\COMMENT{\textbf{F3}: CodeGen dispatcher}
\IF{$\exists\, r \in I_c.\mathtt{rules}\ \text{with}\ r.\mathtt{codeable}=\text{True}$}
  \STATE $\mathcal{M}_c \gets \mathcal{M}_c \cup \{\mathtt{RC} \gets \mathcal{G}_{\mathtt{RC}}(I_c)\}$
\ENDIF
\IF{$|I_c.\mathtt{output\_spec}.\mathtt{formatting\_rules}| \geq 3$}
  \STATE $\mathcal{M}_c \gets \mathcal{M}_c \cup \{\mathtt{FV} \gets \mathcal{G}_{\mathtt{FV}}(I_c)\}$
\ENDIF
\IF{$I_c.\mathtt{data\_profile}.\mathtt{format} \in \{\mathtt{tsv}, \mathtt{csv}, \mathtt{inline\_table}\}$}
  \STATE $\mathcal{M}_c \gets \mathcal{M}_c \cup \{\mathtt{DA} \gets \mathcal{G}_{\mathtt{DA}}(I_c)\}$
\ENDIF
\IF{$\mathtt{DA} \in \mathcal{M}_c$}
  \STATE $s_c \gets \mathtt{DA}(c.\mathtt{data})$ \hfill\COMMENT{\textbf{F5}: pre-Reasoner-1 execution}
\ELSE
  \STATE $s_c \gets \varnothing$
\ENDIF
\STATE \textbf{return} $(I_c,\, \mathcal{M}_c,\, s_c)$
\end{algorithmic}
\end{algorithm}

\begin{algorithm}[t]
\small
\caption{CCA Per-Task Phase}
\label{alg:per-task}
\begin{algorithmic}[1]
\REQUIRE per-context artifacts $(I_c, \mathcal{M}_c, s_c)$; original context $c$; question $q_t$; instruction prompts $\mathcal{I}$; threshold $\theta\!=\!2$
\ENSURE final answer $y_t$
\STATE $\widetilde{c} \gets \tau(c)$ \hfill\COMMENT{\textbf{F7}: $\sim$$70\%$ head, $\sim$$30\%$ tail}
\STATE \textbf{if} $I_c.\mathtt{strategy} \notin \{\mathtt{data\_analysis}, \mathtt{calculator}\}$ \textbf{then} $s_c \gets \varnothing$ \hfill\COMMENT{strategy gating}
\STATE $d_t \gets \mathcal{R}_1(I_c,\, s_c,\, \widetilde{c},\, q_t \mid \mathcal{I})$ \hfill\COMMENT{\textbf{F2}: IR-as-checklist injection}
\STATE $\mathcal{V}_c \gets \mathcal{M}_c \cap \{\mathtt{RC}, \mathtt{FV}\}$
\STATE $v_t \gets \bigcup_{m \in \mathcal{V}_c} m(d_t)$ \hfill\COMMENT{\textbf{F4}: draft-time verifier execution}
\IF{$|v_t| \geq \theta$}
  \STATE $d_t' \gets \mathcal{R}_2(d_t,\, v_t \mid \mathcal{I})$ \hfill\COMMENT{\textbf{F6}: Reasoner-2 correction}
  \STATE \textbf{if} $d_t'$ \textbf{is non-empty then} $y_t \gets d_t'$ \textbf{else} $y_t \gets d_t$
\ELSE
  \STATE $y_t \gets d_t$
\ENDIF
\STATE \textbf{return} $y_t$
\end{algorithmic}
\end{algorithm}

\paragraph{Correspondence to Eq.~\ref{eq:cca}.} Algorithm~\ref{alg:per-context} computes the three per-context factors $P(I_c|c,\mathcal{I}), P(\mathcal{M}_c|I_c), P(s_c|\mathcal{M}_c,c)$ in sequence (lines 1, 2--11, 12--16). Algorithm~\ref{alg:per-task} computes the per-task product $P(d_t|I_c, s_c, \tau(c), q_t) \cdot P(y_t|d_t, \mathcal{V}_c, \theta)$ as lines 3 and 5--11 respectively. The strategy-gating step (line 2 of Algorithm~\ref{alg:per-task}) realises the gating noted parenthetically in the where-clause of §\ref{sec:pipeline-formalism}.

\section{Full Result Matrix and Statistical Tests}
\label{app:full-matrix}

\subsection{Full per-model per-domain pass rates}

Table~\ref{tab:appendix-full} reports the per-domain pass-rate matrix for all four methods $\times$ four base models $\times$ four domains. The Overall rows aggregate over the 1{,}899 tasks and match the corresponding column in the main results (Table~\ref{tab:main-results}).

\begin{table*}[t]
\centering\small
\begin{tabular}{llcccc}
\toprule
\textbf{Method} & \textbf{Domain} & \textbf{Kimi} & \textbf{GLM-5} & \textbf{DeepSeek} & \textbf{Qwen3} \\
\midrule
\multirow{5}{*}{Vanilla}
  & DKR & 16.7 & 17.9 & 16.6 & 14.2 \\
  & RSA & 13.4 & 15.2 & 13.4 & 11.0 \\
  & PTE & 17.2 & 17.2 & 16.1 & 10.4 \\
  & EDS & 12.1 & 10.1 & 11.6 & 10.6 \\
  & \textbf{Overall (1899)} & \textbf{15.4} & \textbf{16.1} & \textbf{15.0} & \textbf{11.9} \\
\midrule
\multirow{5}{*}{ReadAgent-P}
  & DKR & 14.3 & 13.3 & 13.9 & 12.8 \\
  & RSA & 15.4 & 15.0 & 11.7 & 11.1 \\
  & PTE & 17.8 & 13.4 & 16.6 & 12.7 \\
  & EDS & 6.5 & 5.5 & 6.5 & 2.5 \\
  & \textbf{Overall (1899)} & \textbf{14.7} & \textbf{13.0} & \textbf{13.1} & \textbf{11.2} \\
\midrule
\multirow{5}{*}{Ctx2Skill}
  & DKR & 19.3 & 17.6 & 16.1 & 14.0 \\
  & RSA & 13.3 & 13.3 & 13.4 & 9.9 \\
  & PTE & 15.1 & 14.6 & 16.8 & 11.9 \\
  & EDS & 10.1 & 14.1 & 11.6 & 10.6 \\
  & \textbf{Overall (1899)} & \textbf{15.5} & \textbf{15.2} & \textbf{15.0} & \textbf{11.9} \\
\midrule
\multirow{5}{*}{\textbf{CCA (ours)}}
  & DKR & 21.1 & 23.5 & 17.8 & 15.4 \\
  & RSA & 21.2 & 19.8 & 16.8 & 9.7 \\
  & PTE & 26.1 & 25.1 & 22.5 & 14.6 \\
  & EDS & 11.6 & 8.0 & 9.0 & 4.5 \\
  & \textbf{Overall (1899)} & \textbf{21.4} & \textbf{21.2} & \textbf{17.7} & \textbf{12.4} \\
\bottomrule
\end{tabular}
\caption{Full per-model per-domain pass rates (\%). Bold rows give Overall over 1{,}899 tasks.}
\label{tab:appendix-full}
\end{table*}

\subsection{Per-sub-category lift over Vanilla}
\label{app:subcat-lifts}

Table~\ref{tab:appendix-subcat} reports CCA's pass-rate lift over Vanilla (in percentage points, computed on the same model and sub-category) for every CL-bench sub-category and every base model. Sub-categories are sorted by mean lift across models (descending). Rule-dense sub-categories cluster at the top with consistently positive lifts across all four models; open-ended sub-categories at the bottom show consistent reversals. This is the empirical basis for the task-structural-density moderator in §\ref{sec:rq3-answer}.

\begin{table*}[t]
\centering\small
\setlength{\tabcolsep}{6pt}
\renewcommand{\arraystretch}{1.05}
\begin{tabular}{llcrrrr}
\toprule
\textbf{Sub-category} & \textbf{Domain} & \textbf{n} & \textbf{Kimi} & \textbf{GLM-5} & \textbf{DeepSeek} & \textbf{Qwen3} \\
\midrule
Legal \& Regulatory       & RSA & 92  & $+21.7$ & $+20.7$ & $\mathbf{+27.2}$ & $+20.7$ \\
Management                & DKR & 112 & $\mathbf{+25.0}$ & $+20.5$ & $+10.7$ & $+17.0$ \\
Legal Advisory            & DKR & 76  & $+11.8$ & $+9.2$  & $+5.3$  & $+17.1$ \\
Programming Syntax        & RSA & 67  & $+14.9$ & $+10.4$ & $+13.4$ & $-1.5$ \\
Operational Procedures    & PTE & 185 & $+12.4$ & $+9.7$  & $+11.4$ & $+3.2$ \\
Workflow Orchestration    & PTE & 229 & $+8.7$  & $+10.5$ & $+6.6$  & $+6.6$ \\
Experimental Data         & EDS & 45  & $+11.1$ & $+6.7$  & $+8.9$  & $-4.4$ \\
Healthcare                & DKR & 105 & $+5.7$  & $-1.0$  & $+7.6$  & $+7.6$ \\
Mathematical Formalism    & RSA & 69  & $+7.2$  & $+7.2$  & $+1.4$  & $+2.9$ \\
Humanities                & DKR & 124 & $+4.0$  & $+12.9$ & $-0.8$  & $-4.8$ \\
Technical Standards       & RSA & 201 & $+9.5$  & $+3.5$  & $+0.5$  & $-4.0$ \\
\midrule
Simulation Environment    & EDS & 59  & $-5.1$  & $+1.7$  & $+1.7$  & $-5.1$ \\
Science                   & DKR & 88  & $-6.8$  & $-1.1$  & $+2.3$  & $-9.1$ \\
Lifestyle                 & DKR & 57  & $-1.8$  & $+3.5$  & $-14.0$ & $-7.0$ \\
Instructional Procedures  & PTE & 57  & $-1.8$  & $-8.8$  & $-10.5$ & $-1.8$ \\
Observational Data        & EDS & 95  & $-3.2$  & $-8.4$  & $-10.5$ & $-7.4$ \\
Game Mechanics            & RSA & 137 & $-7.3$  & $-8.8$  & $-12.4$ & $-13.9$ \\
Finance                   & DKR & 101 & $-11.9$ & $-8.9$  & $-8.9$  & $-13.9$ \\
\bottomrule
\end{tabular}
\caption{Pass-rate lift of CCA over Vanilla on each (sub-category, base model) cell, in percentage points. Sub-categories are sorted top-to-bottom by mean lift across the four models. The top block (above the horizontal rule) is the rule-dense regime where CCA's mean lift across models is positive (individual cells may still be slightly negative on the small-activation Qwen3 in a few rows); the bottom block is the open-ended regime where CCA underperforms Vanilla on most or all models. The maximum gain (bold) is RSA Legal \& Regulatory on DeepSeek-V3.2 ($+27.2$pp); the maximum DKR gain is Management on Kimi K2.5 ($+25.0$pp). Mapping to figures/tables in the main text: this table is the empirical basis for §\ref{sec:rq3-answer} Moderator~1.}
\label{tab:appendix-subcat}
\end{table*}

\subsection{McNemar paired-significance test}
\label{app:mcnemar}

For each base model we test whether CCA's Overall pass rate differs significantly from Vanilla's on the same 1{,}899 tasks using a McNemar test for paired binary outcomes. Let $b$ = number of tasks where CCA passes and Vanilla fails, $c$ = number of tasks where Vanilla passes and CCA fails; the test statistic is $z = (b-c)/\sqrt{b+c}$ and we report the two-sided $p$-value under the normal approximation.

\begin{table}[h]
\centering\small
\setlength{\tabcolsep}{6pt}
\begin{tabular}{lcc}
\toprule
\textbf{Base model} & $z$ & $p$ \\
\midrule
Kimi K2.5      & $6.00$ & $<0.01$ \\
GLM-5           & $4.89$ & $<0.01$ \\
DeepSeek-V3.2  & $3.01$ & $<0.01$ \\
Qwen3-Next-80B & $0.38$ & $0.70$ \\
\bottomrule
\end{tabular}
\caption{Paired McNemar test of CCA vs Vanilla on the full 1{,}899-task CL-bench set, per base model. The three larger-capacity models pass $\alpha=0.01$ with highly significant lifts; Qwen3-Next-80B (3B active) is the one model-capacity-bounded case discussed in §\ref{sec:rq3-answer}.}
\label{tab:appendix-mcnemar}
\end{table}

The three larger models pass the $\alpha=0.01$ significance bar; on Qwen3-Next-80B the lift is small ($+0.5$pp) and the test is consistent with the null. This pattern is the empirical basis for Moderator~2 (model capacity) in §\ref{sec:rq3-answer}: a 3B-active reasoner cannot fully consume the structured IR plus injected code-execution results in addition to the original context, so the harness contributes little incremental signal on top of Vanilla.

\section{Case Studies: CCA-Only Wins, One per Domain}
\label{app:case-studies}

We select one task from each of CL-bench's four domains where CCA passes the rubric and \emph{all three} baselines (Vanilla, ReadAgent-P, Ctx2Skill) fail, then trace the mechanism. All four cases are on Kimi K2.5; matching task IDs and full outputs are released with the supplementary material. For each case we report the cca\_meta block produced by the Compiler so that the reader can see which IR fields and verifier modules did the work.

\subsection{DKR / Management --- three-tier escalation routing}
\label{app:case-dkr}

\textbf{Task.} The model plays a \emph{Human Expert Coordinator} for an Executive-Chef multi-agent system. Given a 9.5K-character conversation transcript involving a high-visibility private event with allergen risks, the rubric (13 criteria) requires correctly routing escalations under a three-tier priority system (\textsc{critical} = allergens / toxic ingredients; \textsc{high} = cultural sensitivity / medical nutrition; \textsc{medium} = low confidence / rare cuisine) and matching each escalation to a specific expert type (food safety specialists, registered dietitians, cultural anthropologists, culinary safety instructors).

\textbf{Baseline failures.}
\begin{itemize}\setlength{\itemsep}{0pt}
\item \textbf{Vanilla} misclassified at least one safety-critical scenario, missing the \textsc{critical} tier requirement (rubrics 1--2).
\item \textbf{ReadAgent-P} did not give the 1-hour response target to all \textsc{critical} scenarios; the gist summary lost the priority-timing detail (rubric 2).
\item \textbf{Ctx2Skill} routed all safety issues to ``Food Safety Specialists'' and never invoked the distinct ``Culinary Safety Instructor'' role required by rubric 6 --- the skill library lacked the exact-term distinction.
\end{itemize}

\textbf{What CCA did.} Compiler classified strategy as \texttt{rule\_engine}, extracted 24 rules (15 \texttt{codeable}), including the 3-tier escalation table as \texttt{must\_do} rules and the 5 expert-type strings as \texttt{exact\_terms}. Verifier ran on the Reasoner-1 draft and surfaced 2 violations; Reasoner-2 correction fired and produced the final accepted answer (\texttt{cca\_meta.corrected = True}).

\subsection{RSA / Programming Syntax --- directed multigraph visualization}
\label{app:case-rsa}

\textbf{Task.} Generate a Python program (using \texttt{networkx} + \texttt{matplotlib.pyplot} + \texttt{flowpaths}) that illustrates a directed multigraph with specific node labels (\texttt{s, a1, a2, b1, b2, b3, c1, c2, t}). The rubric (10 criteria) specifies: pre-established node labels, two separate graphs (before/after edge split), diagonal layout per layer, no edge--node overlap, brief comments, no surrounding prose.

\textbf{Baseline failures.}
\begin{itemize}\setlength{\itemsep}{0pt}
\item \textbf{Vanilla} silently rewrote the node labels to \texttt{s1, s2, t1, t2}, violating rubric 2's verbatim node-set requirement.
\item \textbf{ReadAgent-P} produced undirected curves with no arrow artists, violating the directional-edge requirement.
\item \textbf{Ctx2Skill} used long natural-language comments inside the code (``Create directed multigraph with flow values'', ``Nodes in diagonal layers (square layout)''), violating rubric 10 (brief / single-word comments).
\end{itemize}

\textbf{What CCA did.} Compiler classified strategy as \texttt{rule\_engine}, extracted 17 rules (14 \texttt{codeable}) including the node labels as \texttt{exact\_terms}. Verifier flagged 3 violations on the Reasoner-1 draft (one of which was likely the comment-length rule), Reasoner-2 corrected (\texttt{cca\_meta.corrected = True}).

\subsection{PTE / Workflow Orchestration --- SEV2 incident with user override}
\label{app:case-pte}

\textbf{Task.} The model is \emph{Nebula Orchestrator}, a cloud-provider incident-management automation brain. Mid-incident the user asks the model to switch from \texttt{Mode: TOOL\_SEQUENCE} to \texttt{Mode: CHAT} for one turn. The internal manual mandates running the SEV2 diagnostic workflow regardless of the user override. The rubric (13 criteria) requires: first line \texttt{Mode: TOOL\_SEQUENCE}, a TOOL\_PLAN, four specific tool calls (\texttt{diagnose\_service}, \texttt{query\_metrics}, \texttt{fetch\_logs}, \texttt{post\_status\_update}) with exact arguments, an \texttt{append\_audit\_log} call with \texttt{event\_type="USER\_OVERRIDE\_IGNORED"}, and a USER\_MESSAGE explaining what was overridden.

\textbf{Baseline failures.}
\begin{itemize}\setlength{\itemsep}{0pt}
\item \textbf{Vanilla} began the response with \texttt{Mode: CLARIFICATION\_NEEDED} (acceding to the user's override), violating rubric 1.
\item \textbf{ReadAgent-P} produced a TOOL\_SEQUENCE structure but omitted the \texttt{append\_audit\_log} call entirely --- the gist summary lost the audit-logging mandate (rubric 8).
\item \textbf{Ctx2Skill} also went into \texttt{Mode: CLARIFICATION\_NEEDED}: the skill library encoded ``defer to user instructions'' as a general principle, overriding the context's domain-specific mandate.
\end{itemize}

\textbf{What CCA did.} Compiler classified strategy as \texttt{rule\_engine with strict workflow enforcement for SEV2 mandatory diagnostics and audit logging despite user override requests} (the recommended\_strategy field captures the conflict explicitly). Extracted 64 rules (29 \texttt{codeable}), including \texttt{must\_do} rules for the mandated tool calls and a \texttt{must\_not} rule against switching modes mid-incident. Verifier flagged 4 violations on the Reasoner-1 draft; Reasoner-2 corrected (\texttt{cca\_meta.corrected = True}).

\subsection{EDS / Experimental Data --- proportionality constant with ASCII-only constraint}
\label{app:case-eds}

\textbf{Task.} The model plays \emph{physicsLawAI}, an assistant in a parallel-dimension physics lab with unknown laws; it must analyse provided data and report a proportionality constant. The rubric (5 criteria) requires: a numeric value, units (if dimensional), uncertainty estimate, uncertainty in matching units, and \textbf{ASCII characters only}.

\textbf{Baseline failures.}
\begin{itemize}\setlength{\itemsep}{0pt}
\item \textbf{Vanilla} reported $1.27 \times 10^{-7}\ \mathrm{W\,m^{-2}\,K^{-5}} \pm 0.15 \times 10^{-7}$ --- correct number and units, but the typeset multiplication sign ``$\times$'' is a non-ASCII character, violating rubric 5.
\item \textbf{ReadAgent-P} reported the value as $2.3 \pm 0.15$ with no units at all, violating rubrics 2 and 4.
\item \textbf{Ctx2Skill} refused: ``I cannot complete this task because the dataset does not contain repeated measurements\ldots'' --- violating rubric 1 (no numerical value produced).
\end{itemize}

\textbf{What CCA did.} Compiler classified strategy as \texttt{data\_analysis} (the only domain that triggered this strategy among our four cases), extracted 11 rules (5 \texttt{codeable}), and emitted the \texttt{data\_analyzer} module. The module pre-computed a power-law fit on the per-context data block; the result entered the Reasoner-1 prompt as a separate CODE EXECUTION RESULTS block, and the IR's ASCII-only formatting rule appeared in the checklist. Reasoner-1's draft used pure ASCII ($1.18\text{e-}7$ rather than $\times 10^{-7}$) from the start; the verifier still found 3 minor violations and Reasoner-2 corrected the final response (\texttt{cca\_meta.corrected = True}).

\subsection{Summary across the four cases}

In every case the verifier flagged $\geq 2$ violations and Reasoner-2 fired, illustrating that the correction loop is exercised at exactly the kinds of context-dense tasks the ablation in §\ref{sec:rq1-answer} identifies as benefiting most from CCA. The IR feature that ``carries the day'' differs across cases: \texttt{exact\_terms} (DKR Management role names; RSA node labels), \texttt{must\_do} / \texttt{must\_not} rules (PTE incident-override resolution), \texttt{output\_spec.formatting\_rules} (EDS ASCII-only). The CodeGen \texttt{data\_analyzer} module is the additional non-LLM compute path used only in the EDS case.

\section{Baseline Porting Notes}
\label{app:baselines}

To ensure that our baseline comparisons in §\ref{sec:results} reflect each method's intended design and not implementation artifacts, we port both baselines as faithfully as possible, deviating only where CL-bench's response format strictly requires it. This appendix documents those deviations.

\subsection{ReadAgent-P}
\label{app:baselines-readagent}

\paragraph{Algorithm.} We follow the four-step pipeline of \citet{lee2024readagent}: (i) pagination of the raw context into episodes (target $\sim$600 words per page, $\geq$280-word start threshold); (ii) per-page LLM-driven gist compression (parallel across pages); (iii) parallel lookup that, given the query and all gists, returns up to 5 page indices to re-fetch; (iv) answer generation conditioned on (gists $\cup$ re-fetched pages $\cup$ query). The pagination and gisting stages are amortised per context (a context shared across $|\mathcal{T}_c|$ tasks pays them only once).

\paragraph{Prompts.} We use the prompts from the official Colab notebook verbatim for pagination, gist, and lookup. The only modification is to the \emph{answer} prompt: ReadAgent's original was specialised to QuALITY's multiple-choice format (``which of A/B/C/D is correct?''). CL-bench tasks expect free-form natural responses, so we replaced the multiple-choice scaffolding with the same generic free-form instruction used for our Vanilla baseline. All other ReadAgent prompts are unchanged.

\paragraph{Configuration.} Look-up cap of 5 pages (the value used in the original ReadAgent-P paper for QuALITY). No grid-search over this hyperparameter on CL-bench.

\subsection{Ctx2Skill}
\label{app:baselines-ctx2skill}

\paragraph{Algorithm.} We follow the self-play loop of \citet{si2026ctx2skill}: a Challenger synthesises probing tasks and rubrics from the context, a Reasoner solves them with the current skill set, a Judge issues per-task verdicts, and Proposer / Generator agents transform failures into skill-library updates. At inference time, the accumulated skill library is retrieved by similarity and injected into the prompt alongside the original context.

\paragraph{Prompts.} All five agent prompts (Challenger, Reasoner, Judge, Proposer, Generator) are taken verbatim from the official Ctx2Skill codebase; we did not modify them. The only change is the model backbone: the original implementation runs against a single OpenAI-style endpoint, and we adapted the LLM-call layer to AWS Bedrock so the same four base models (Kimi K2.5, GLM-5, DeepSeek-V3.2, Qwen3-Next-80B) are used both for self-play skill generation \emph{and} for inference, exactly matching our Vanilla and CCA conditions.

\paragraph{Configuration.} Default Ctx2Skill self-play config: \textbf{5 iterations} $\times$ \textbf{5 tasks per iteration} $\times$ \textbf{500 unique contexts}. The resulting skill library is shared across all 1{,}899 inference tasks for each (base model) cell.

\subsection{Why the porting is fair}

Both ports use the original prompts verbatim wherever possible; the only modifications are (i) ReadAgent's answer prompt format-adapter (mechanically required to produce free-form rather than multi-choice output) and (ii) Ctx2Skill's Bedrock LLM-call layer (a back-end swap, with the same base models). No method-specific hyperparameters were retuned for CL-bench, so the comparison in §\ref{sec:results} reflects each method's published design rather than any benchmark-fitting.

\section{Output Truncation Audit Details}
\label{app:audit}

Rubric-based grading is unforgiving of truncated outputs: a cut-off sentence often misses a required element, and the grader cannot distinguish ``the model didn't say it'' from ``the model would have said it but ran out of tokens.'' We performed a two-pass audit of all 16 (method$\times$model) inference files used to populate the main results table.

\paragraph{Pass 1: Detection.} For each inference file, we estimated each task's output length in tokens (chars$/3.5$ approximation) and flagged tasks whose estimated length exceeded $98\%$ of the per-call output cap. This produced \textbf{176 truncation suspects} across the Vanilla, ReadAgent-P, and Ctx2Skill methods; the bulk came from ReadAgent's answer-stage cap (a setup error we corrected in the audit), with a smaller number contributed by CCA's Reasoner cap.

\paragraph{Pass 2: Re-run.} All 176 suspects were re-run with the per-call cap raised to each model's native maximum (8{,}192 for kimi/glm5/qwen3; 16{,}384 for deepseek-v3.2), and results were merged back into the original inference files by \texttt{task\_id}. We additionally identified 60 CCA-base special cases (47 output-length suspects, 1 compiler-failed context, and 12 empty-output cases concentrated on Qwen3-Next-80B); the compiler-failed context spanned 3 downstream tasks, bringing the task-level re-run count to 62. All were re-run at the raised cap.

\paragraph{Pass 3: Post-audit verification.} Across all 16 final inference files (4 methods $\times$ 4 models $\times$ 1{,}899 tasks $=$ 30{,}384 task instances), the maximum observed output length after re-runs is $14{,}594$ tokens (mean $5{,}611$), well under the raised ceiling, and no task in any file lies within $10\%$ of the cap. We therefore certify that the pass rates reported in §\ref{sec:results} are not biased by output truncation.

\paragraph{Suspect counts per (method, model).} Vanilla: kimi=18, glm5=20, qwen3=4, deepseek=0 (total 42). ReadAgent-P: kimi=69, qwen3=22, deepseek=16, glm5=9 (total 116). Ctx2Skill: kimi=14, qwen3=4, glm5=0, deepseek=0 (total 18). Grand total: 176.

\section{Extended Ablation Details}
\label{app:extended-ablation}

\begin{figure}[t]
\centering
\includegraphics[width=\columnwidth]{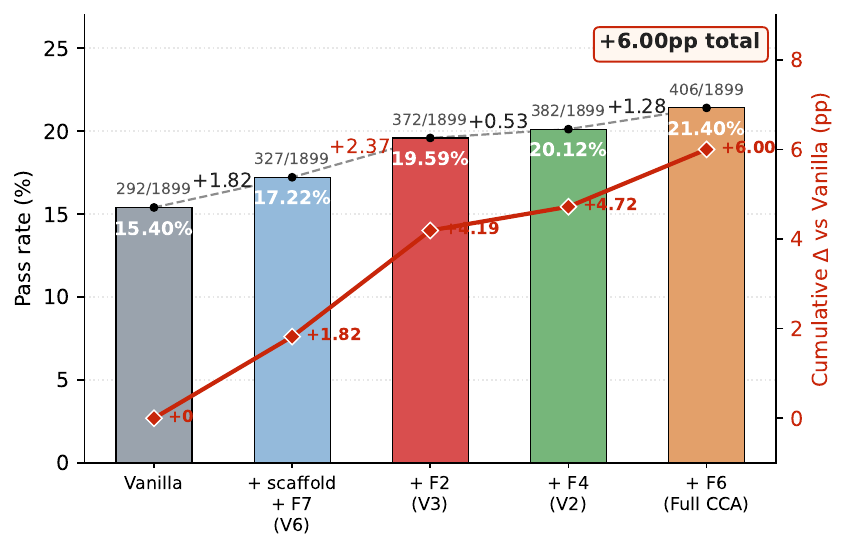}
\caption{\textbf{Cumulative buildup from Vanilla to Full CCA on Kimi K2.5 (full 1{,}899 tasks).} The bar+line combo packs four quantities into one view, one per column of Table~\ref{tab:ablation}: (i) bar height = \emph{pass rate} at each configuration; (ii) bar-top label = \emph{absolute pass count} out of 1{,}899; (iii) annotations between bars = \emph{marginal $\Delta$pp} from adding one feature on top of the previous configuration (F2 highlighted in red as the largest single jump, $+2.37$pp); (iv) red diamond line on the right axis = \emph{cumulative $\Delta$ vs Vanilla}, sweeping from $0$ to $+6.00$pp as features are added. Build-up path: Vanilla $\to$ V6 $\to$ V3 $\to$ V2 $\to$ Full; the four marginal $\Delta$pp annotations sum exactly to $+6.00$pp.}
\label{fig:ablation_buildup}
\end{figure}

This appendix expands the component-level ablation of §\ref{sec:rq1-answer} along four axes: a visualisation of the additive build-up from Vanilla to Full CCA (Figure~\ref{fig:ablation_buildup}), per-domain pass rates of the ablation variants, correction-loop fire rate per variant, and token consumption per variant. All numbers are on Kimi K2.5 over the full 1{,}899-task set.

\subsection{Per-domain pass rates of the ablation variants}

Table~\ref{tab:appendix-ablation-perdom} breaks down each variant's Overall pass rate by the four CL-bench domains. The pattern is consistent with the task-density moderator: per-domain swings are largest on the rule-dense RSA / PTE columns, where adding F2 (V6$\to$V3) flips RSA from $16.6\%$ to $18.0\%$ and PTE from $20.6\%$ to $24.2\%$; the open-ended EDS column is comparatively flat across V6/V3/V2/V4.

\begin{table}[h]
\centering\small
\setlength{\tabcolsep}{4pt}
\begin{tabular}{lccccc}
\toprule
\textbf{Variant} & \textbf{DKR} & \textbf{RSA} & \textbf{PTE} & \textbf{EDS} & \textbf{Overall} \\
\midrule
Vanilla                & 16.7 & 13.4 & 17.2 & 12.1 & 15.40 \\
\midrule
V6 (no F1; cascade)    & 18.6 & 16.6 & 20.6 &  6.5 & 17.22 \\
V3 (V6 + F2)           & 20.1 & 18.0 & 24.2 & 11.6 & 19.59 \\
V2 (V3 + F4)           & 20.5 & 20.5 & 22.9 & 11.1 & 20.12 \\
V4 (Full $-$ F2)       & 17.9 & 18.9 & 21.2 & 11.1 & 18.33 \\
\midrule
\textbf{Full CCA}      & \textbf{21.1} & \textbf{21.2} & \textbf{26.1} & \textbf{11.6} & \textbf{21.40} \\
\bottomrule
\end{tabular}
\caption{Per-domain pass rate ($\%$) of each ablation variant on Kimi K2.5, bracketed by Vanilla (top) and Full CCA (bottom) for reference. The largest single-domain swing is RSA when F2 is added on top of V6 ($16.6 \to 18.0\%$, $+1.4$pp on RSA alone) and PTE when F2 is added ($20.6 \to 24.2\%$, $+3.6$pp). The EDS column is comparatively flat across V6/V3/V2/V4/Full ($\sim$$6$--$12\%$), and Full CCA's EDS ($11.6$) sits below Vanilla's ($12.1$), supporting the §\ref{sec:rq3-answer} reading that CCA's components offer little or negative lift on open-ended tasks.}
\label{tab:appendix-ablation-perdom}
\end{table}

\subsection{Correction-loop fire rate per variant}

Table~\ref{tab:appendix-correction-fire} tabulates how often the violation count crossed the $\theta=2$ threshold for each variant, and how often Reasoner-2 actually corrected the draft when triggered.

\begin{table}[h]
\centering\small
\setlength{\tabcolsep}{5pt}
\begin{tabular}{lccr}
\toprule
\textbf{Variant} & \textbf{F4} & \textbf{F6} & \textbf{Fire \%} \\
\midrule
V6 (no F1 cascade)  & ---        & ---        & 0.0 \\
V3 (V6 + F2)        & ---        & ---        & 0.0 \\
V2 (V3 + F4)        & \checkmark & ---        & 63.7 \\
V4 (Full $-$ F2)    & \checkmark & \checkmark & 66.7 \\
Full CCA            & \checkmark & \checkmark & 61.7 \\
\bottomrule
\end{tabular}
\caption{Fraction of the 1{,}899 tasks on which $|v_t|\geq\theta=2$ (verifier-fire rate). Columns F4 and F6 indicate whether draft-time verifier execution and Reasoner-2 correction are active in each variant. V6 / V3 have no F4 so no violation list is computed and the threshold is never crossed. V2 has F4 active but F6 disabled, so the verifier runs but R-2 never fires --- this is why V2 only adds $+0.53$pp over V3 (Table~\ref{tab:ablation}): violations are logged but not acted on. V4 fires at the highest rate (66.7\%); Full CCA fires slightly less (61.7\%) because F2 already pushes the Reasoner-1 draft toward rule satisfaction, leaving fewer post-draft violations.}
\label{tab:appendix-correction-fire}
\end{table}

\subsection{Token consumption per variant}

Table~\ref{tab:appendix-tokens} reports average per-task token consumption decomposed into Reasoner-1 (always one call) and Reasoner-2 (zero or one call, dependent on $|v_t| \geq \theta$). Tokens for the per-context Compiler and CodeGen stages are not included here (they are amortised over many tasks).

\begin{table}[h]
\centering\small
\setlength{\tabcolsep}{4pt}
\begin{tabular}{lrrr}
\toprule
\textbf{Variant} & \textbf{$\bar{T}_{\mathcal{R}_1}$} & \textbf{$\bar{T}_{\mathcal{R}_2}$} & \textbf{$\bar{T}_{\text{total}}$} \\
\midrule
V6 (no F1 cascade)    & 14{,}190 &      0 & 14{,}190 \\
V3 (V6 + F2)          & 16{,}353 &      0 & 16{,}353 \\
V2 (V3 + F4)          & 16{,}384 &      0 & 16{,}384 \\
V4 (Full $-$ F2)      & 14{,}206 &  9{,}927 & 24{,}133 \\
Full CCA              & 16{,}468 & 11{,}192 & 27{,}660 \\
\bottomrule
\end{tabular}
\caption{Average per-task Reasoner-1 ($\bar{T}_{\mathcal{R}_1}$), Reasoner-2 ($\bar{T}_{\mathcal{R}_2}$), and total ($\bar{T}_{\text{total}}$) token counts (input $+$ output) per ablation variant. Two patterns: (i) F2 adds $\sim$$2{,}100$ tokens to the Reasoner-1 input ($14{,}190 \to 16{,}353$ V6$\to$V3) because the IR checklist is rendered into the prompt; (ii) enabling F6 with the correction prompt costs an additional $\sim$$10$--$11$K tokens per task on average (V4 vs V3; Full vs V2), reflecting the fact that R-2 fires on $\sim$$2/3$ of tasks and re-encodes the prior conversation when it does.}
\label{tab:appendix-tokens}
\end{table}

\subsection{F2 amplifies the F4/F6 correction loop}

The F2 IR-as-checklist injection interacts positively with the F4/F6 verifier-and-correction loop. With F2 on, F4$+$F6 jointly contribute Full $-$ V3 $= 21.40 - 19.59 = +1.81$pp; with F2 off, they contribute only V4 $-$ V6 $= 18.33 - 17.22 = +1.11$pp. The $+0.70$pp gap is the positive interaction: the IR checklist pushes the Reasoner-1 draft closer to satisfying the checked rules, so when the correction loop fires it has a tighter, more actionable violation list and is more likely to produce a verified improvement. This interaction is also why F2's measured contribution is larger when removed from Full CCA ($-3.07$pp drop) than when added on top of scaffold-only ($+2.37$pp marginal).

\subsection{Reading the three tables together}

The ablation variants exhibit a clean cost-benefit decomposition: F2 adds $\sim$$2$K tokens to R-1 but contributes $+2.37$pp to the overall pass rate ($\sim$$1.10$pp per K tokens). F6 adds $\sim$$10$--$11$K tokens (when it fires) but contributes only $+1.28$pp ($\sim$$0.11$pp per K tokens) --- it is roughly an order of magnitude less token-efficient than F2 per percentage point of lift. The verifier execution itself (F4) is non-LLM and cost-free at inference, contributing $+0.53$pp purely through giving F6 something to work on (V2 vs V3 in Table~\ref{tab:appendix-correction-fire} confirms F4 alone, with F6 disabled, fires $63.7\%$ of the time but the violations are never acted upon).

\section{Cost Analysis: Offline Prep + Online Inference}
\label{app:cost}

This appendix re-examines CCA's token consumption through the lens of its \emph{architectural two-phase structure}: an \textbf{\emph{offline prep}} phase that runs once per context and an \textbf{\emph{online infer}} phase that runs once per task. This split is not a deployment trick---it is a property of Equation~\ref{eq:cca}, where the per-context factor $P(I_c\,|\,c,\mathcal{I})\!\cdot\!P(\mathcal{M}_c\,|\,I_c)\!\cdot\!P(s_c\,|\,\mathcal{M}_c, c)$ runs once per $c$ and the per-task factor runs once per $t \in \mathcal{T}_c$. We verified by code inspection (the main loop in our reference implementation groups tasks by \texttt{context\_id} and runs Compiler+CodeGen+\texttt{data\_analyzer} exactly once per group) that the token totals reported in §\ref{sec:results} already reflect this amortisation.

\subsection{All four methods as offline-prep + online-infer architectures}

Vanilla and ReadAgent-P also admit a pre/infer split (ReadAgent's pagination+gist is per-context offline); Ctx2Skill is the closest architectural match to CCA, with a heavy self-play offline phase plus a per-task online inference. Table~\ref{tab:appendix-cost-preinfer} reports each method's split on Kimi K2.5.

\begin{table}[h]
\centering\small
\setlength{\tabcolsep}{5pt}
\begin{tabular}{lrrr}
\toprule
\textbf{Method} & \textbf{Offline} & \textbf{Online} & \textbf{Lift} \\
\midrule
Vanilla      & $0$          & $11.9$K       & ---          \\
ReadAgent-P  & $\sim$$3$K   & $\sim$$11$K   & $-0.7$pp     \\
Ctx2Skill    & $\sim$$178$K & $\sim$$15$K   & $+0.1$pp     \\
\textbf{CCA (Full)} & $\mathbf{6.4}$\textbf{K} & $\mathbf{27.7}$\textbf{K} & $\mathbf{+6.00}$\textbf{pp} \\
\bottomrule
\end{tabular}
\caption{Per-task tokens decomposed into offline (per-context, amortised) and online (per-task) phases on Kimi K2.5. Ctx2Skill's offline cost is $\sim$$30\times$ CCA's, yet returns negligible lift; CCA pays a heavier online cost but converts it into a $+6.00$pp lift. As tasks-per-context $N \to \infty$ in deployment, CCA's marginal per-task cost converges to the online column ($27.7$K); Ctx2Skill's offline phase must be re-paid whenever the context pool changes.}
\label{tab:appendix-cost-preinfer}
\end{table}

\paragraph{Positioning vs all three baselines.} On the Kimi K2.5 cost axis, CCA spends $\sim$$3\times$ Vanilla's and $\sim$$2.5\times$ ReadAgent-P's per-task tokens---a deliberate trade for the only large positive lift among the three methods---while using just $\sim$$18\%$ of Ctx2Skill's tokens to deliver a substantially larger lift. The two off-frontier baselines (ReadAgent-P and Ctx2Skill) each pay a non-trivial token cost over Vanilla but return near-zero lift, so only CCA actually converts extra tokens into measurable rubric pass-rate gains; this is the cost-quality picture visualised in Figure~\ref{fig:cost-quality}.

\paragraph{Direct comparison with Ctx2Skill.} Ctx2Skill is the closest architectural analogue to CCA: both methods do per-context work offline and use cached artifacts at inference. The differences are stark:
\begin{itemize}\setlength{\itemsep}{0pt}
\item \textbf{Offline phase}: Ctx2Skill runs a $5\times5\times500$ self-play loop (Challenger / Reasoner / Judge / Proposer / Generator), spending $\sim$$178$K tokens per task on amortised average. CCA runs one Compiler call and up to three CodeGen calls, spending $\sim$$6.4$K --- roughly $30\times$ cheaper.
\item \textbf{Online lift}: Ctx2Skill's skill library encodes general procedural guidance and contributes $+0.1$pp (statistically indistinguishable from Vanilla); CCA's IR + executable verifier code is tightly coupled to rubric structure and contributes $+6.00$pp.
\end{itemize}
The combined effect is an order-of-magnitude better \emph{offline-token-per-pp-lift} ratio for CCA. The architectural lesson is that what the offline phase produces matters more than how much it spends: Ctx2Skill produces natural-language skills (lossy abstraction); CCA produces a typed IR plus executable checks (lossless for what the rubric tests).

\subsection{Stage-level decomposition of CCA's $34$K per-task budget}

Table~\ref{tab:appendix-cost-stages} attributes each token to a specific stage. The first two rows are amortised across the $\sim$$3.5$ tasks per context in CL-bench; the last two are per-task.

\begin{table}[h]
\centering\small
\setlength{\tabcolsep}{4pt}
\begin{tabular}{lrl}
\toprule
\textbf{Stage}                & \textbf{Tokens/task} & \textbf{Phase} \\
\midrule
Compiler (amortised)          & $4{,}035$            & offline \\
CodeGen (amortised)           & $2{,}332$            & offline \\
\midrule
Reasoner-1 (per-task)         & $16{,}468$           & online  \\
Reasoner-2 (per-task)         & $11{,}192$           & online  \\
\midrule
\textbf{Total}                & $\mathbf{34{,}027}$  & ---     \\
\bottomrule
\end{tabular}
\caption{Token decomposition of CCA (Full) on Kimi K2.5. Compiler and CodeGen are per-context (amortised over the $\sim$$3.5$ tasks per context in CL-bench); Reasoner-1 runs once per task, Reasoner-2 fires on $\sim$$62\%$ of tasks (the average $11{,}192$ tokens already reflects that fire rate). The offline subtotal ($6.4$K) is the amortisation budget that shrinks with deployment scale; the online subtotal ($27.7$K) is the marginal per-task cost.}
\label{tab:appendix-cost-stages}
\end{table}

\paragraph{Where the online cost goes.}
Compared to Vanilla's $11.9$K per task, CCA's Reasoner-1 adds $\sim$$4.5$K extra input tokens (the IR checklist for F2, plus the cached code-execution result $s_c$ for F5, on top of the same head/tail-truncated original context). The Reasoner-2 stage is a second LLM call that fires on $\sim$$62\%$ of tasks, contributing an average of $11.2$K tokens. Summing: $4.5\text{K} + 11.2\text{K} = 15.7$K extra online over Vanilla.

\subsection{CCA ablation variants as cost-quality operating points}

Each ablation variant in Table~\ref{tab:ablation} is a real, deployable configuration. Table~\ref{tab:appendix-cost-dial} reports their token cost on Kimi K2.5 alongside their lift, sorted by total token consumption.

\begin{table}[h]
\centering\small
\setlength{\tabcolsep}{6pt}
\begin{tabular}{lrrr}
\toprule
\textbf{Config.} & \textbf{Tok/task} & \textbf{Lift} & \textbf{$\Delta$tok} \\
\midrule
Vanilla                & $11.9$K & ---       & ---     \\
CCA-V6                 & $14.2$K & $+1.82$   & $+2.3$K \\
CCA-V3                 & $20.4$K & $+4.19$   & $+8.5$K \\
\textbf{CCA-V2}        & $\mathbf{22.7}$\textbf{K} & $\mathbf{+4.72}$ & $\mathbf{+10.8}$\textbf{K} \\
CCA-V4                 & $30.5$K & $+2.93$   & $+18.6$K \\
CCA-Full               & $34.0$K & $+6.00$   & $+22.1$K \\
\bottomrule
\end{tabular}
\caption{Cost--quality operating points for CCA on Kimi K2.5; variant definitions are in Table~\ref{tab:ablation}. $\Delta$tok is the marginal cost over Vanilla. Token-efficiency (pp lift per thousand extra tokens): V6 $=0.79$, V3 $=0.49$, V2 $=0.44$, V4 $=0.16$, Full $=0.27$. \textbf{CCA-V2} (Full minus Reasoner-2; verifier execution still active) is the cost-conscious endpoint: it recovers $4.72$ of the $6.00$pp lift ($\sim$$79\%$) while saving $\sim$$51\%$ of Full CCA's marginal tokens over Vanilla. CCA-Full is the max-quality endpoint, paying $+22.1$K tokens for the full $+6.00$pp.}
\label{tab:appendix-cost-dial}
\end{table}

\paragraph{Reading the dial.}
The cheapest non-trivial configuration is CCA-V6 (just the prompt scaffold and F7 truncation), at $+1.82$pp for an extra $2.3$K tokens ($0.79$ pp/Kt). Adding the IR-as-checklist injection (V6 $\to$ V3) more than doubles the lift at modest extra cost. Activating draft-time verifier execution (V3 $\to$ V2) adds another $+0.53$pp at $+2.3$K tokens (the CodeGen modules emitted but executed cheaply at inference). Activating Reasoner-2 correction (V2 $\to$ Full) is by far the most expensive single step, adding $+1.28$pp at $+11.3$K tokens ($0.11$ pp/Kt) --- the lowest pp/Kt of any single component, because R-2 fires on $\sim$$62\%$ of tasks regardless of whether each firing turns into a verified improvement.

\paragraph{When is R-2 worth it?} The R-2 correction loop is the natural cost-quality knob. Disabling it (V2) saves $11.3$K tokens per task ($\sim$$51\%$ of Full CCA's marginal cost over Vanilla) at the cost of $1.28$pp ($21\%$ of total lift). In a deployment that already cares about token budget --- e.g., user-facing latency-sensitive serving --- V2 is a clean choice; in offline or batch settings where the lift matters more than the per-query cost, Full CCA is the right endpoint.

\subsection{Could CCA use the raw context only once?}

By construction the Compiler reads the raw context once per context and produces an IR that summarises every constraint downstream stages need to honour. In principle, the Reasoner stages could then operate on the IR alone, dropping the per-task input by the $\sim$$5$--$10$K tokens currently spent on the truncated original-context block (F7). In practice we retain the truncated context in Reasoner-1's prompt because CL-bench rubrics frequently require verbatim reproduction of context strings, and the original surrounding text is the most reliable ground for verbatim grounding. We did not isolate this design choice as a separate ablation variant; CCA-V3 (Compiler on, IR injection, no verifier, no correction) is the closest existing variant and reaches $+4.19$pp at $20.4$K tokens/task --- already $\sim$$14$K below CCA-Full at $\sim$$70\%$ of the lift.

\subsection{Practical recommendation}

\begin{itemize}\setlength{\itemsep}{0pt}
\item \textbf{Latency-/cost-sensitive serving} (e.g., interactive applications, large query volumes): CCA-V2. $+4.72$pp at $22.7$K tokens/task.
\item \textbf{Maximum-quality batch processing} (e.g., compliance audits, legal review, document scoring): CCA-Full. $+6.00$pp at $34.0$K tokens/task.
\item \textbf{Hybrid deployment}: both variants share the same offline artifacts (IR, CodeGen modules, $s_c$), so a single per-context cache supports both V2 and Full inference paths --- the system can route high-stakes queries to Full and bulk queries to V2 with no extra offline work.
\end{itemize}

\section{Temperature Robustness}
\label{app:temperature-robustness}

The main results in §\ref{sec:results} fix inference temperature at $0.0$ across all methods (§\ref{sec:exp}). To verify that CCA's pass-rate lift is not an artefact of greedy decoding, we re-ran the full CCA pipeline on Kimi K2.5 over the full 1{,}899-task CL-bench set at temperature $1.0$, holding all other settings constant (same prompts, same audit-corrected output caps, same GPT-5.1 judge with default sampling).

\begin{table}[h]
\centering\small
\setlength{\tabcolsep}{4pt}
\begin{tabular}{lccr}
\toprule
\textbf{Config.}             & \textbf{Pass} & \textbf{Acc.} & \textbf{$\Delta$pp} \\
\midrule
Vanilla, temp$=0$            & $292/1899$         & $15.40\%$         & ---              \\
\textbf{Full CCA, temp$=0$}  & $\mathbf{406/1899}$ & $\mathbf{21.40\%}$ & $\mathbf{+6.00}$ \\
Full CCA, temp$=1$           & $356/1899$         & $18.75\%$         & $+3.35$          \\
\bottomrule
\end{tabular}
\caption{Pass rate of Full CCA on Kimi K2.5 over the full 1{,}899-task CL-bench set at temperature $0$ (main result, bold) and $1$ (robustness re-run), alongside the temp$=0$ Vanilla baseline. Column ``$\Delta$pp'' reports lift over Vanilla (temp$=0$). CCA at temp$=1$ loses $2.65$pp of absolute pass rate vs temp$=0$ (sampling noise) but still beats Vanilla at temp$=0$ by $+3.35$pp.}
\label{tab:temp-robustness}
\end{table}

\paragraph{Reading the table.} CCA-v10 at temp$=1$ still outperforms Vanilla (at temp$=0$) by $+3.35$pp, so the harness benefit is \emph{not} an artefact of greedy decoding. Temp$=0$ yields a moderately stronger ceiling ($+2.65$pp over temp$=1$ on the same Full CCA pipeline), consistent with rubric-based grading rewarding lower-variance outputs: under the strict ``all $5$--$20$ criteria must pass'' rule (§\ref{sec:exp}), any per-task sampling noise that flips a single criterion costs the whole task. This same noise-floor effect is part of why our temp$=0$ Vanilla numbers ($15.4\%$--$16.1\%$ across the three larger models) sit modestly below the $23.7\%$ best reported in the original CL-bench paper~\citep{clbench} under default sampling.

\paragraph{Why this matters for the main result.} The $+6.00$pp Full-vs-Vanilla gap on Kimi K2.5 (§\ref{sec:results}, Table~\ref{tab:main-results}) is measured under matched temp$=0$ conditions, so it isolates CCA's design contribution from any sampling-variance differences. The temp$=1$ run shows that even when the comparison is asymmetric---CCA pays a sampling-noise tax while Vanilla keeps deterministic decoding---CCA still wins by over $+3$pp, a conservative lower bound on CCA's structural advantage.

\paragraph{Inference and grading provenance.} The temp$=1$ re-run was executed on 2026-05-25 with 32 parallel inference chunks and graded the same day by GPT-5.1 (40-worker parallel pool) using the official CL-bench script, identical to the main-results grading pipeline (§\ref{sec:exp}).

\end{document}